\documentclass{article}

\usepackage{libs/arxiv}

\usepackage[acronym]{glossaries}
\usepackage[inline]{enumitem}
\usepackage[numbers]{natbib}
\usepackage{hyperref}
\usepackage{mathtools}
\usepackage{chngcntr}
\usepackage{booktabs}
\usepackage{amsmath}
\usepackage{import}
\usepackage{xspace}
\usepackage{array}
\usepackage{tikz}

\glsdisablehyper
\newcommand{\TITLE}{Persistent Animal Identification Leveraging Non-Visual Markers}

\newcommand{\repo}{\url{https://github.com/michael-camilleri/TIDe}}

\newlength{\pgsize}
\newcommand{\wrt}{w.r.t.\@\xspace}

\newcommand{\Naive}{Na\"\i ve\xspace}

\newcommand{\eg}{e.g.\@\xspace}

\newcommand{\ie}{i.e.\@\xspace}

\newcommand{\etal}{{\em et al}.\@\xspace}
\newcommand{\vs}{vs.\@\xspace}

\newcommand{\sect}{Sect.\@\xspace}

\newcommand{\acta}{Actual Analytics\textsuperscript{\texttrademark}\@\xspace}

\newcommand{\supp}{Supplementary Material\@\xspace}

\newcommand{\Prob}[1]{\ensuremath{P\left(#1\right)}}
\newcommand{\Transpose}{\ensuremath{^\top}}

\newcommand{\STracklets}{\ensuremath{\mathbf{S}}}
\newcommand{\OTracklet}{\ensuremath{s}}

\newcommand{\OObject}{\ensuremath{o}}
\newcommand{\OPosition}{\ensuremath{p}}
\newcommand{\ITime}{\ensuremath{t}}
\newcommand{\MMseTrackWeight}{\ensuremath{w}}

\newacronym{fnr}{FNR}{false negative rate}
\newacronym{fpr}{FPR}{false positive rate}
\newacronym{fps}{FPS}{frames per second}

\newacronym{hca}{HCA}{Home Cage Analyser}

\newacronym{jpda}{JPDA}{Joint Probabilistic Data Association}

\newacronym{ias}{IAS}{Identity Assignment System}
\newacronym{ilp}{ILP}{Integer Linear Programming}
\newacronym{impc}{IMPC}{International Mouse Phenotype Consortium}
\newacronym{iou}{IoU}{Intersection-over-Union}
\newacronym{ir}{IR}{Infra-Red}

\newacronym{kf}{KF}{Kalman Filter}

\newacronym[longplural={Bounding Boxes},shortplural={BBs}]{bb}{BB}{Bounding Box}
\newacronym{lr}{LR}{Linear Regression}

\newacronym{map}{MAP}{Maximum-A-Posteriori}
\newacronym{mnap}{mAP}{Mean Average Precision}
\newacronym{mot}{MOT}{Multi-Object Tracking}
\newacronym{mota}{MOTA}{Multi-Object Tracking Accuracy}
\newacronym{motp}{MOTP}{Multi-Object Tracking Precision}
\newacronym{mrc}{MLC at MRC Harwell}{Mary Lyon Centre at MRC Harwell, Oxfordshire}

\newacronym{nb}{NB}{\Naive Bayes}
\newacronym{nn}{NN}{Neural Network}

\newacronym{reid}{re-ID}{re-identification}
\newacronym{rf}{RF}{Random Forest}
\newacronym{rfid}{RFID}{Radio-Frequency Identification}

\newacronym{sota}{SOTA}{state-of-the-art}

\usetikzlibrary{shapes}
\usetikzlibrary{fit}
\usetikzlibrary{chains}
\usetikzlibrary{arrows}

\tikzstyle{latent} = [circle,fill=white,draw=black,inner sep=2pt,
minimum size=30pt, font=\fontsize{10}{10}\selectfont, node distance=1]
\tikzstyle{latent_big} = [latent, minimum size=40pt]
\tikzstyle{latent_med} = [latent, minimum size=35pt]
\tikzstyle{latent_sml} = [latent, minimum size=20pt, font=\fontsize{8}{8}\selectfont]
\tikzstyle{obs} = [latent,fill=gray!25]
\tikzstyle{obs_big} = [obs, minimum size=45pt]
\tikzstyle{obs_med} = [obs, minimum size=35pt]
\tikzstyle{obs_sml} = [obs, minimum size=20pt, font=\fontsize{8}{8}\selectfont]
\tikzstyle{const_lat} = [rectangle, draw=black, inner sep=2pt, node distance=1, fill=white, minimum size=15pt]
\tikzstyle{const} = [rectangle, draw=black, inner sep=2pt, node distance=1, fill=gray!25, minimum size=15pt]
\tikzstyle{blank} = [rectangle, inner sep=0pt, node distance=1]
\tikzstyle{factor} = [rectangle, fill=black, minimum size=5pt, inner
sep=0pt, node distance=0.4]
\tikzstyle{det} = [latent, diamond]

\tikzstyle{plate} = [draw, rectangle, rounded corners, fit=#1, inner sep=5pt]
\tikzstyle{wrap} = [inner sep=0pt, fit=#1]
\tikzstyle{gate} = [draw, rectangle, dashed, fit=#1]

\tikzstyle{caption} = [font=\footnotesize, node distance=0] %
\tikzstyle{plate caption} = [caption, node distance=0, inner sep=0pt,
right=5pt of #1.south east] %
\tikzstyle{factor caption} = [caption] %
\tikzstyle{every label} += [caption] %

\newcommand{\edge}[3][]{ %
  \foreach \x in {#2} { %
    \foreach \y in {#3} { %
      \path (\x) edge [->, >={triangle 45}, #1] (\y) ;%
    } ;
  } ;
}

\newcommand{\plate}[4][]{ %
  \node[wrap=#3] (#2-wrap) {}; %
  \node[plate caption=#2-wrap] (#2-caption) {#4}; %
  \node[plate=(#2-wrap)(#2-caption), #1] (#2) {}; %
}

\title{\TITLE}
\date{}

\renewcommand{\headeright}{Preprint}
\newcommand{\sout}[1]{}

\author{
  Michael P. J. Camilleri \\
  School of Informatics,\\
  University of Edinburgh \\
   \And
  Li Zhang \\
  School of Data Science,\\
  Fudan University
   \And
  Rasneer S. Bains \\
  Mary Lyon Centre, \\
  MRC Harwell
   \And
  Andrew Zisserman \\
  Dept. of Engineering Science,\\
  University of Oxford
   \And
  Christopher K. I. Williams \\
  School of Informatics\\
  University of Edinburgh\\
}

\newcommand{\parencite}[1]{\cite{#1}}

\newcommand\blfootnote[1]{%
\begingroup
\renewcommand\thefootnote{}\footnote{#1}%
\addtocounter{footnote}{-1}%
\endgroup
}

\begin{document}

\maketitle

% Abstract
\begin{abstract}
Our objective is to locate and provide a unique identifier for each mouse in a cluttered home-cage environment through time, as a precursor to automated behaviour recognition for biological research.
This is a very challenging problem due to (i) the lack of distinguishing visual features for each mouse, and (ii) the close confines of the scene with constant occlusion, making standard visual tracking approaches unusable.
However, a coarse estimate of each mouse's location is available from a unique \acrshort{rfid} implant, so there is the potential to optimally combine information from (weak) tracking with coarse information on identity.

To achieve our objective, we make the following key contributions: (a) the formulation of the \emph{object identification} problem as an assignment problem (solved using \acrlong{ilp}), (b) a novel probabilistic model of the affinity between tracklets and \acrshort{rfid} data, and (c) a curated dataset with per-frame \glsfirst{bb} and regularly-spaced ground-truth annotations for evaluating the models.
The latter is a crucial part of the model, as it provides a principled probabilistic treatment of object detections given coarse localisation.
Our approach achieves 77\% accuracy on this animal identification problem, and is able to reject spurious detections when the animals are hidden.
\end{abstract}

\blfootnote{\small\noindent This version of the article has been accepted for publication, after peer review but is not the Version of Record and does not reflect post-acceptance improvements, or any corrections.
The Version of Record is available online at: \url{https://doi.org/10.1007/s00138-023-01414-1}.}

% Rest of Chapters
\section{Introduction}
\label{S_INTRO}

\begin{figure}
	\centering
	\includegraphics[width=0.95\textwidth]{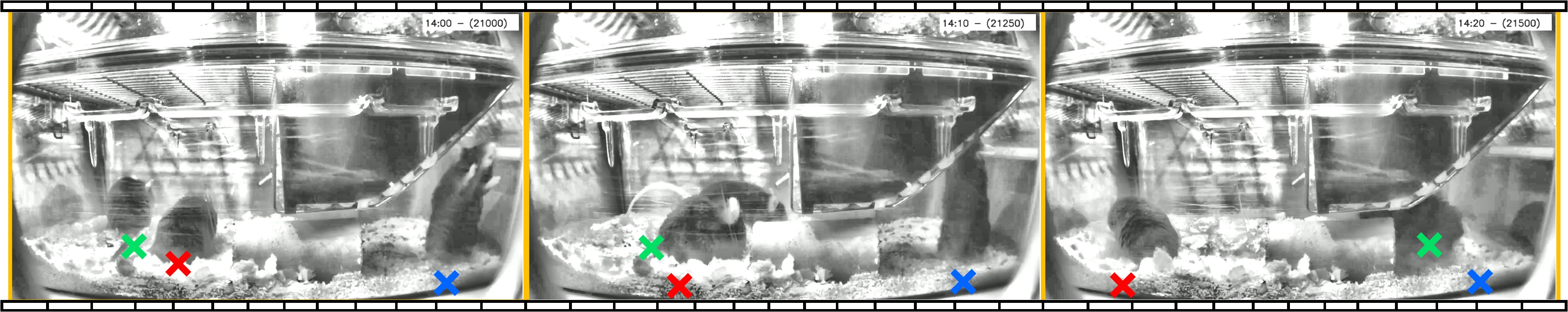}
  	\caption{Sample frames from our dataset. To improve visualisation, the frames are processed with CLAHE \cite{MISC_038} and brightened: our methods operate on the raw frames. The mice are visually indistinguishable, their identity being inferred through the \gls{rfid} pickup, shown as Red/Green/Blue crosses projected into image-space. This is enough to distinguish the mice when they are well separated (left): however, as they move around, they are occluded by cage elements (Green in centre) or by cage mates (Green by Blue in right) and we have to reason about temporal continuity and occlusion dynamics.}
  	\label{FIG_DATASET_EXAMPLE}
\end{figure}

We are motivated by the problem of tracking and identifying group-housed mice in videos of a cluttered home-cage environment, as in Fig.\ \ref{FIG_DATASET_EXAMPLE}.
This \emph{animal identification} problem goes beyond tracking to assigning unique identities to each animal, as a precursor to automatically annotating their individual behaviour (\eg, feeding, grooming) in the video, and analysing their interactions.
It is also distinct from object classification (recognition) \cite{DS_014}: the mice have no distinguishing visual features and cannot be merely treated as different objects.
Their visual similarity, as well as the close confines of the enriched cage environment make standard visual tracking approaches very difficult, especially when the mice are huddled or interacting together.
We found experimentally that standard trackers alone break down into outputting short tracklets, with spurious and missing detections, and thus do not provide a persistent identity.
However, we can leverage additional information provided by a unique \gls{rfid} implant in each mouse, to provide a coarse estimate of its location (on a $3 \times 6$ grid).
This setting is not unique to animal tracking and can be applied to more general objects.
Similar situations arise \eg, when identifying specific team-mates in robotic soccer \cite{MISC_052} (where the object identification is provided by the weak self-localisation of each robot), or identifying vehicles observed by traffic cameras at a busy junction (as required for example for law-enforcement), making use of a weak location signal provided by cell-phone data.

Our solution starts from training and running a mouse detector on each frame, and grouping the detections together with a tracker \cite{VL_051} to produce a set of tracklets, as shown in Fig.\ \ref{FIG_ARCHITECTURE}(c).
We then formulate an assignment problem to \emph{identify} each tracklet as belonging to a mouse (based on \gls{rfid} data) or a dummy object (to capture spurious tracklets).
This problem is solved using \gls{ilp}, see Fig.\ \ref{FIG_ARCHITECTURE}(f).
Tracking is used to reduce the complexity of the problem relative to a frame-wise approach and implicitly enforce temporal continuity for the solution.

The problem setup is not standard \gls{mot}, since we go beyond tracking by \emph{identifying} individual tracklets using additional \gls{rfid} information which provides persistent unique identifiers for each animal.
Our main contribution is thus to formulate this \emph{object identification} as an assignment problem, and solve it using \gls{ilp}.
A key part of the model is a principled probabilistic treatment for modelling detections from the coarse localization information, which is used to provide assignment weights to the \gls{ilp}.

We emphasize that we are solving a real-world problem which is of considerable importance to the biological community (as per the \gls{impc} \cite{PHT_013}), and which needs to scale up to thousands of hours of data in an efficient manner --- this necessarily affects some of our design choices.
More generally, with Home-Cage monitoring systems being more readily available to scientists \parencite{MS_004}, there are serious efforts in maximising their use\footnote{See \eg the TEATIME consortium \url{https://www.cost-teatime.org/}}. 
Consequently there is a real need for analysis methods like ours, particularly where multiple animals are co-housed.

In this paper we first frame our problem in light of recent efforts (\sect \ref{S_RELATED_WORK}), indicating how our situation is quite unique in its formulation.
We discuss our approach from a theoretical perspective in \sect \ref{S_METHODS}, postponing the implementation mechanics to \sect \ref{S_IMPLEMENTATION}.
Finally, we showcase the utility of our solution through rigorous experiments (\sect \ref{S_EXPERIMENTS}) on our dataset which allows us to analyse its merits in some depth.
A more detailed description of the dataset, details for replicating our experimental setups (including parameter fine-tuning) as well as deeper theoretical insights are relegated to the \supp.
The curated dataset, together with code and some video clips of our framework in action are available at \url{https://github.com/michael-camilleri/TIDe} (see \sect\ref{S_CONCLUSION} and Appendix A in the Supplementary Document for details).

\section{Related work}
\label{S_RELATED_WORK}
Below we discuss related work \wrt \gls{mot}, \gls{reid} and animal tracking. 
We defer discussion of our \gls{ilp} formulation relative to other work after discussing the method in \sect \ref{SS_IDENTIFICATION}.

\paragraph{\acrlong{mot}:}
There have been many recent advances in \gls{mot} \cite{VL_009}, fuelled in part by the rise of specialised neural architectures \cite{VL_028, VL_049, VL_063}.
Here the main challenge is keeping track of an undefined number of \emph{visually distinct} objects (often people, as in \eg \cite{DS_004, DS_020}), over a short period of time (as they enter and exit a particular scene) \cite{DS_018}.
However, our problem is not \gls{mot} because we need to \emph{identify} (rather than just track) a fixed set of individuals using persistent, externally-assigned identities, rather than relative ones (as in \eg \cite{VL_065}): \ie our problem is not indifferent to label-switching.
We care about the absolute identity assigned to each tracklet -- even if we had input from a perfect tracker, there would still need to be a way to assign the anonymous tracklets to identities, as we do below.

There is another important difference between the current \acrlong{sota} in \gls{mot} (see \eg \cite{VL_008, VL_049, VL_040, VL_062, VL_065, VL_015, VL_012}) and our use-case.
As the mice are essentially indistinguishable, we cannot leverage appearance information to distinguish them but must rely on external cues -- the \gls{rfid}-based quantised position.
On top of this, there are also the challenges of operating in a constrained cage environment (which exacerbates the level of occlusion), and the need to consistently and efficiently track mice over an extended period of time (on the order of hours).

\paragraph{(Re-)Identification:} 
\Gls{reid} \cite{VL_067} addresses the transitory notion of identity by using appearance cues to join together instances of the same object across `viewpoints'.
Obviously, this does not work when objects are visually indistinguishable, but there is also another key difference.
The standard \gls{reid} setup deals with associating together instances of the same object, often (but not necessarily) across multiple non-overlapping cameras/viewpoints \cite{VL_066}.
In our specific setup, however, we wish to relate objects to an external identity (\eg \gls{rfid}-based position); our work thus sits on top of any algorithm that builds tracklets (such as that of Fleuret \etal \cite{VL_067}).

\paragraph{Animal Tracking:}
Although generally applicable, this work was conceived through a collaboration with the Mary Lyon Centre at MRC Harwell \parencite{CBD_026}, and hence seeks to solve a very practical problem: analysis of group-housed mice in their enriched home-cage environment over extended periods of 3-day recordings.
This setup is thus significantly more challenging than traditional animal observation models, which often side-step identification by involving single animals in a purpose-built arena \cite{CBD_041, CBD_038, VL_022, VL_050} rather than our enriched home-cage.
Recent work on multiple-animal tracking \cite{VL_040, VL_056, DS_027, DS_028} often uses visual features for identification which we cannot leverage (our subjects are visually identical):
for \eg the CalMS21 \cite{DS_027} dataset uses recordings of pairs of visually distinct mice, while Marshall \etal \cite{DS_028} attach identifying visual markers to their subjects in the PAIR-R24M dataset.
Most work also requires access to richer sources of colour/depth information \cite{VL_034, VL_036} rather than our single-channel \gls{ir} feed, or employs top-mounted cameras \cite{VL_036, VL_037, VL_032,VL_035, VL_040} which give a much less cluttered view of the animals.
Indeed, systems such as idtracker.ai \cite{VL_055} or DeepLabCut \cite{VL_059, VL_069} do not work for our setup, since the mice often hide each other and keypoints on the animals (which are an integral part of the methods) are not consistently visible (besides requiring more onerous annotations).
Finally we reiterate that our goal is to \emph{identify} tracklets through fusion with external (\gls{rfid}) information, which none of the existing frameworks support --- for example the multi-animal version of DeepLabCut \parencite{VL_069} only supports supervised identity learning for visually distinct individuals.

\section{Methodology}
\label{S_METHODS}

\begin{figure}
	\centering
	\begin{tabular}{>{\centering\arraybackslash}m{0.55\textwidth}@{\hspace{1em}}>{\centering\arraybackslash}m{0.4\textwidth}}
		\includegraphics[width=0.55\textwidth]{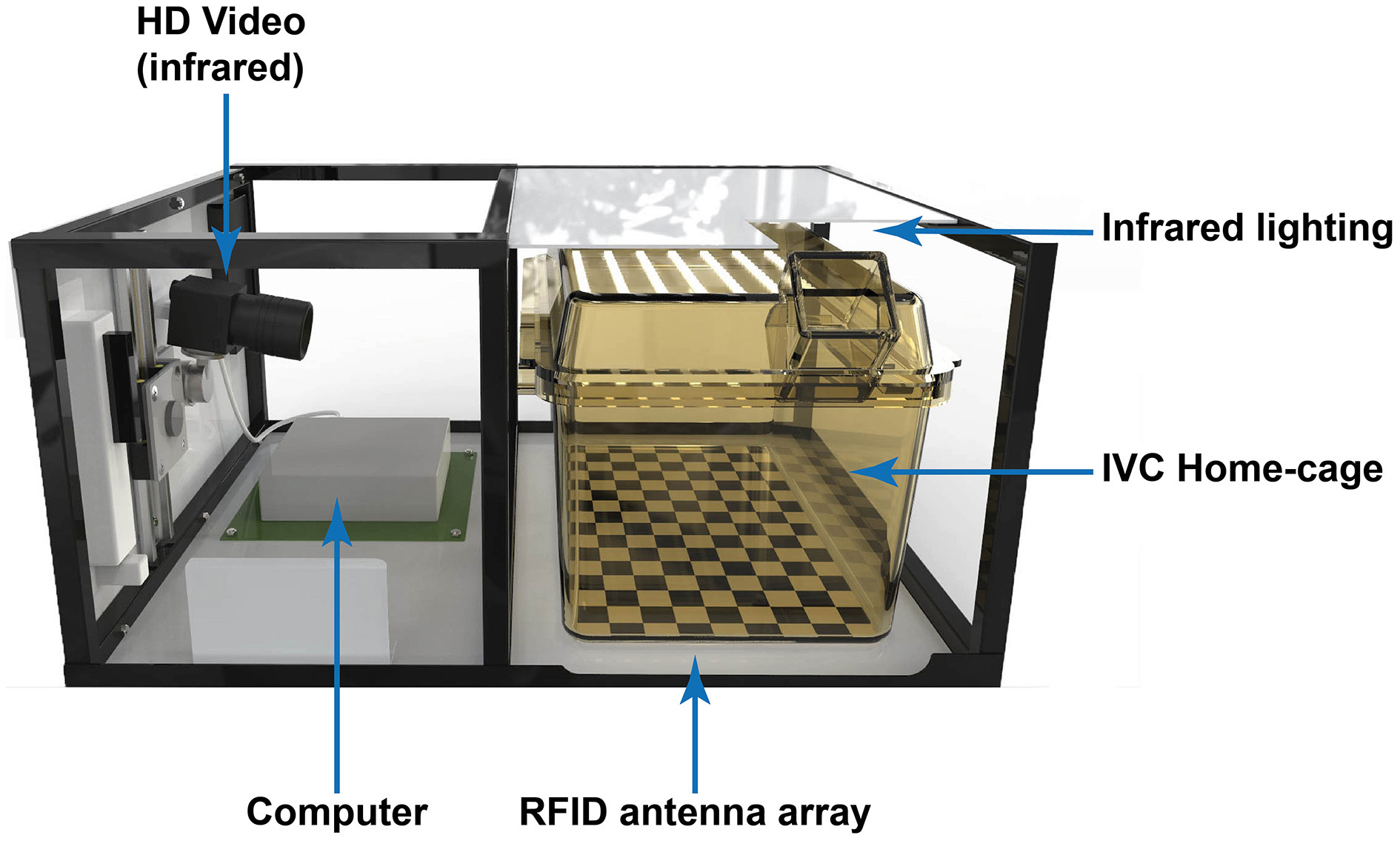} &
		\includegraphics[width=0.4\textwidth]{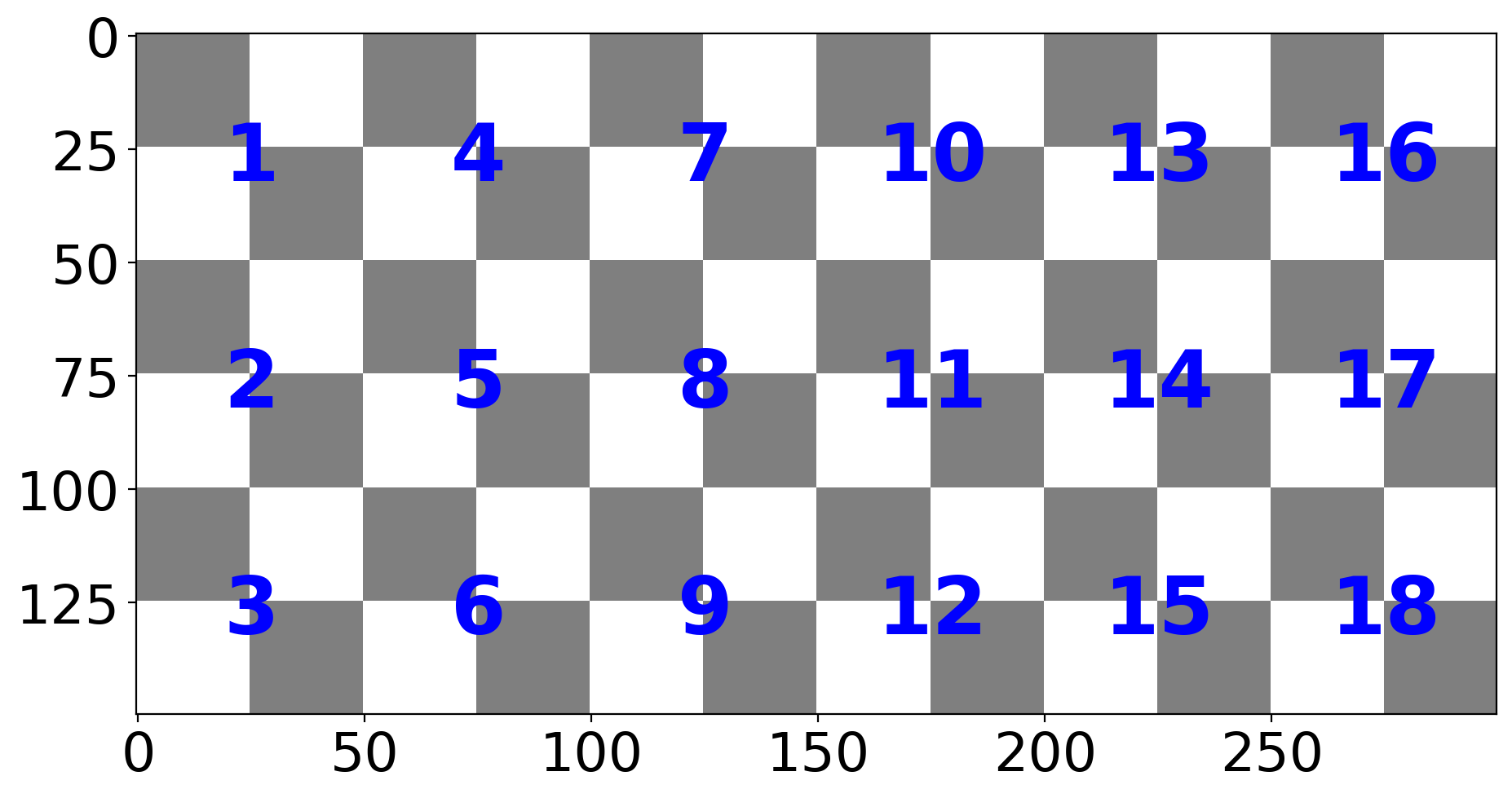} \\
		(a) & (b)
	\end{tabular}
  	\caption{The Actual Analytics Home Cage Analysis system. (a) The rig used to capture mouse recordings (Illustration reproduced with permission from \cite{CBD_026}). (b) The (numbered) points on the checkerboard pattern used for calibration (measurements are in mm): the numbers correspond also to the \gls{rfid} antenna receivers on the baseplate.}
  	\label{FIG_HCA_RIG}
\end{figure}

This section describes our proposed approach to identification of a fixed set of animals in video data.
We explain the methodology through the running example of tracking group-housed mice.
Specifically, consider a scenario in which we have continuous video recordings of mice housed in groups of three as shown in Fig.\ \ref{FIG_DATASET_EXAMPLE}.
The data, captured using a setup similar to that shown in Fig.\ \ref{FIG_HCA_RIG}(a), consists of single-channel \gls{ir} side-view video and coarse \gls{rfid} position pickup from an antenna grid below the cage as in Fig.\ \ref{FIG_HCA_RIG}(b).
Otherwise, the mice have no visual markings for identification.
Implementation details are deferred to \sect \ref{SS_EXP_DATASET}.

\begin{figure*}
	\centering
	\includegraphics[width=0.98\textwidth]{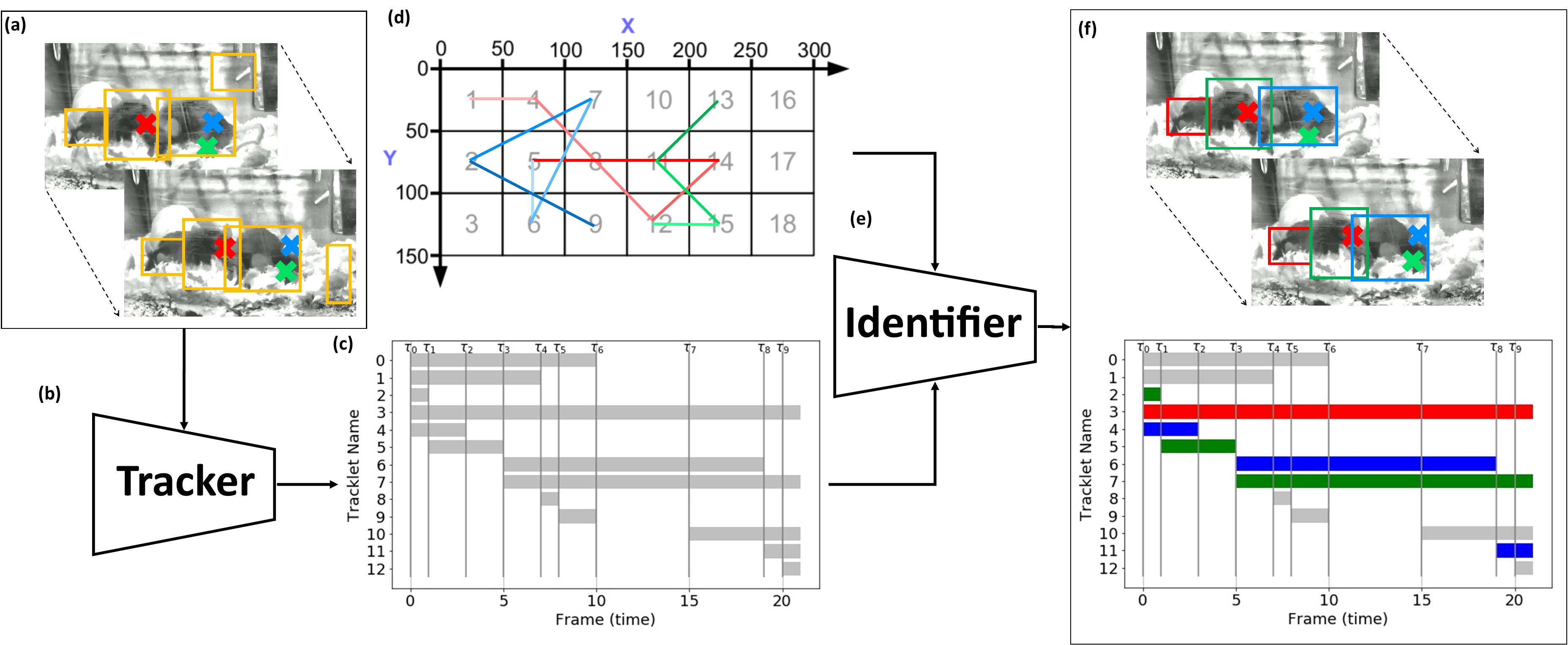}
  	\caption{Our proposed architecture. Given per-frame detections of animals (a), a tracker (b) joins these into tracklets, shown (c) in terms of their temporal span. After discarding spurious detections based on a lifetime threshold, the identifier (e) combines these with coarse localisation traces over time (d), to produce identified \acrfullpl{bb} for each object throughout the video, shown [f] as coloured tracklets/\glspl{bb}.}
  	\label{FIG_ARCHITECTURE}
\end{figure*}

\paragraph{Overview:}
Our system takes in candidate detections about the fixed set of animals in the form of \acrfullpl{bb} and assigns an identity to each of them using a two-stage pipeline, as shown in Fig.\ \ref{FIG_ARCHITECTURE}.
First, we use a \emph{Tracker} \cite{VL_051} to join the detections across frames into tracklets based on optimising the \gls{iou} between \glspl{bb} in adjacent frames.
This stage injects temporal continuity into the object identification, filters out spurious detections, and, as will be seen later, reduces the complexity of the identification problem.
Then, an \emph{Identifier}, using a novel \gls{ilp} formulation, combines the tracklets with the coarse position information to identify which tracklet(s) belong to each of the known animals, based on a probabilistic weight model of object locations.

\subsection{Tracking}
We use the tracker to inject temporal continuity into our problem, encoding the prior that animals can only move a limited distance between successive frames.
Subsequently, we formulate our object identification problem as assigning \emph{tracklets} to animals over a recording interval indexed by $\ITime \in \left\lbrace 1, ..., T\right\rbrace$.

The tracker also serves the purpose of simplifying the optimisation problem.
While \glspl{bb} and positions are registered at the video frame rate, it is not necessary to run our optimisation problem (below) at this granularity.
Rather, we group together the sequence of frames for which the same subset of tracklets are active: a new interval will begin when a new tracklet is spawned, or when an existing one disappears.
The observation is thus broken up into intervals, indexed sequentially by $\ITime$, shown as $\tau_{\lbrace 0, \dots, 9 \rbrace}$ at the top of Fig.\ \ref{FIG_ARCHITECTURE} (c).

\subsection{Object Identification from tracklets}
\label{SS_IDENTIFICATION}
The tracker produces $I$ tracklets: to this we add $T$ special `hidden' tracklets for which no \glspl{bb} exist, and which serve to model occlusion at any time-step. 
The goal of the Identifier is then to assign object identities to the set of tracklets, \STracklets, of size $I+T$.
The subset of tracklets active at time $\ITime$ is represented by $\STracklets^{\lbrace\ITime\rbrace}$.
We assume there are $J$ animals $\OObject_1, \ldots, \OObject_J$ we wish to track/identify (and for which we have access to coarse location through time); an extra dummy object $\OObject_{J+1}$ is a special outlier/background model which captures spurious tracklets in $\STracklets$.

We also define a utility $\MMseTrackWeight_{i j}$, which is a measure of the quality of the match between tracklet $\OTracklet_i$ and animal $\OObject_j$: we assume that tracklets cannot switch identities and are assigned in their entirety to one animal (this is reasonable if we are liberal in breaking tracklets).
The weight is typically a function of the \gls{bb} features and animal position (see \sect \ref{SS_ASSIGNMENT_WEIGHTING}).
Given the above, our \gls{ilp} optimises the assignment matrix $A$ with elements $a_{i j} \in \left\lbrace 0, 1\right\rbrace$ (where $a_{i, j} = 1 \Leftrightarrow$ `$\OTracklet_i$ is assigned to $\OObject_j$'), that maximises:
\begin{align}
\max_{A} & \sum_{i=1}^{I+T} \sum_{j=1}^{J+1} \MMseTrackWeight_{i j} a_{i j} , \label{EQ_TRK_OBJECTIVE}\\
 \shortintertext {\hspace{1cm} subject to the constraints:} & 
 	\sum_{j=1}^{J+1} a_{i j} = 1 \quad \forall\ i \in I , \label{EQ_TRK_LPC_1MSE}\\
 					& \sum_{s \in \STracklets^{\lbrace t \rbrace}} a_{s j} = 1 \quad \forall\ t \in T,\ j \in J , \label{EQ_TRK_LPC_1TRK} \\
 					& a_{i (J+1)} = 0 \quad \forall \  I+1 \leq i \leq I+T .\label{EQ_TRK_INVALID}
\end{align}

Constraint \eqref{EQ_TRK_LPC_1MSE} ensures that a \emph{generated} tracklet is assigned to exactly one animal, which could be the outlier model.
Constraint \eqref{EQ_TRK_LPC_1TRK}, enforces that each animal (excluding the outlier model) is assigned exactly one tracklet (which could be the `hidden' tracklet) at any point in time, and finally, Eq.\ \eqref{EQ_TRK_INVALID} prevents assigning the hidden tracklets to the outlier model.
Unfortunately, these constraints mean that the resulting linear program is not Totally Unimodular \cite{MTH_017}, and hence does not automatically yield integral optima.
Consequently, solving the \gls{ilp} is in general NP-Hard, but we have found that due to our temporal abstraction, using a branch and cut approach efficiently finds the optimal solution on the size of our data without the need for any approximations.

Our \gls{ilp} formulation of the object identification process is related to the general family of covering problems \cite{MTH_007} as we show in our \supp (see \sect B.1).
In this respect, our problem is a generalisation of Exact-Set-Cover \cite{MTH_008} in that we require every animal to be `covered' by one tracklet at each point in time: \ie we use the stronger equality constraint rather than the inequality in the general \gls{ilp} (see Eq.\ (B.2)).
Unlike the Exact-Set-Cover formulation, however, 
\begin{enumerate*}[label=(\alph*)]
     \item \label{IT_SET_MC} we have multiple objects (animals) to be covered, 
     \item \label{IT_SET_OTHERS} some objects need not be covered at all (outlier model), \textsl{and}
     \item \label{IT_SET_ALL} all detected tracklets must be used (although a tracklet can cover the `extra' outlier model).
\end{enumerate*}
Note that with respect to \ref{IT_SET_MC}, this is not the same as Set Multi-Cover \cite{MTH_009} in which the same object must be covered more than once: \ie $b_f$ in Eq.\ (B.2) is always 1 for us.

\subsection{Assignment weighting model}
\label{SS_ASSIGNMENT_WEIGHTING}
The weight matrix with elements $w_{ij}$ is the essential component for incorporating the quality of the match between tracklets and locations.
It is easier to define the utility of assignments on a per-frame basis, aggregating these over the lifetime of the tracklet.
The per-frame weight boils down to the level of agreement between the features of the \gls{bb} and the position of the object at that frame, together with the presence of other occluders.

We can use any generative model for this purpose, with the caveat that we can only condition the observations (\glspl{bb}) on fixed information which does not itself depend on the result of the \gls{ilp} assignment.
Specifically, we can use all the \gls{rfid} information and even the presence of the tunnel, but not the locations of other \glspl{bb}.
The reason for this is that if we condition on the other detections, the weight will depend on the other assignments (\eg whether a \gls{bb} is assigned to a physical object or not), which invalidates the \gls{ilp} formulation.

We define our weight model as the probability that an animal picked up by a particular \gls{rfid} antenna $j$, or the outlier distribution $J+1$, could have generated the \gls{bb} $i$.
For the sake of our \gls{ilp} formulation, we also allow animal $j$ to generate a hidden \gls{bb} when it is not visible.

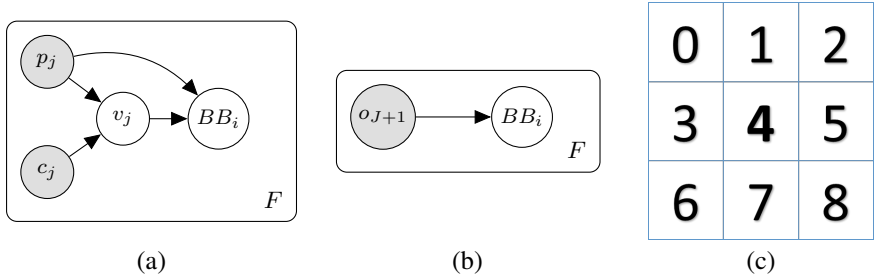
\begin{figure*}
	\centering
	\begin{tabular}{>{\centering\arraybackslash}m{4.2\pgsize}@{\hspace{1em}}>{\centering\arraybackslash}m{3.5\pgsize}@{\hspace{1em}}>{\centering\arraybackslash}m{3.6\pgsize}}
		\begin{tikzpicture}
  \node[obs_sml]            							(Ri)	{$\OPosition_{j}$};
  \node[obs_sml, below=0.75cm of Ri]    					(Ci)	{$c_j$};
  \node[latent_sml, below right=0.25cm and 0.5cm of Ri] (Vi)	{$v_j$};
  \node[latent_sml, right=0.5cm of Vi]  				(Bj)    {$BB_i$};
%  \node[obs_sml, right=1cm of Ci]						(Ro)	{$\OObject_J$};
%  \node[latent_sml, below=1.5cm of Ro]	(Bj2)   {$BB_i$};
  
  \edge {Ri} {Vi}
  \edge[bend left] {Ri} {Bj}
  \edge {Ci} {Vi}
  \edge {Vi} {Bj}
%  \edge {Ro} {Bj2}

  \plate {F} {(Ri)(Ci)(Vi)(Bj)} {$F$}; %
\end{tikzpicture} &
		\begin{tikzpicture} 
  \node[obs_sml]						(Ro)	{$\OObject_{J+1}$};
  \node[latent_sml, right=1cm of Ro]	(Bj2)   {$BB_i$};
  
  \edge {Ro} {Bj2}
  
  \plate {F} {(Ro)(Bj2)} {$F$};
\end{tikzpicture} &
		\includegraphics[width=3cm]{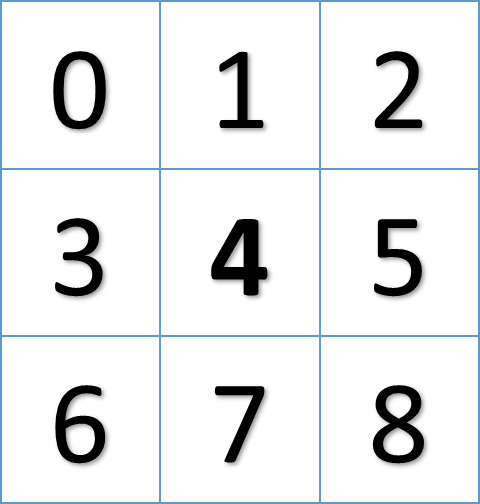} \\
		(a) & (b) & (c)
	\end{tabular}
  	\caption{(a) Model for assigning $BB_i$ to $\OObject_j$ per-frame in $F$ (frame index omitted to reduce clutter) when $\OObject_j$ is an animal. Variable $p_j$ refers to the position of animal $j$, and $c_j$ captures contextual information. The visibility is represented by $v_j$, and the \gls{bb} (actual or occluded) by $BB_i$. (b) Model for assigning $BB_i$ under the outlier model. (c) Neighbourhood definition for representing $c_j$.}
  	\label{FIG_WEIGHT_MODEL}
\end{figure*}

For objects of interest ($j \leq J$), we employ the graphical model in Fig.\ \ref{FIG_WEIGHT_MODEL} (a).
Variable $\OPosition_{j}$ represents the position of object $j$: $c_j$ captures contextual information, which impacts the visibility $v_j$ of the \gls{bb} and is a function of the \gls{rfid} pickups of the other animals in the frame.
The visibility, $v_j$, can take on one of three values: 
\begin{enumerate*}[label=(\arabic*)]
\item clear (fully visible),
\item truncated (partial occlusion), \textsl{and}
\item hidden (object is fully occluded and hence no \gls{bb} appears for it).
\end{enumerate*}
The \gls{bb} parameters are then conditioned on both the object position (for capturing the relative arrangement within the observation area) as well as the visibility as:
\begin{equation}
\Prob{BB_i|p_j, v_j} \sim  \\
	\begin{cases}
	     1 & \text{if } v_j = \text{Hidden \& } i \geq I+1 \\
	     0 & \text{if } v_j = \text{Hidden \& } i < I+1 \\
		 \mathcal{N}\left(\mu_{pv}, \Sigma_{pv}\right) & \text{otherwise}
	\end{cases} , \label{EQ_BB_PROB}
\end{equation}
where the parameters of the Gaussian are in general a function of object position and visibility.
Note that in the top line of Eq.\ \eqref{EQ_BB_PROB} we allow only a hidden object to generate a `hidden' tracklet.
Conversely, under the outlier model, Fig.\ \ref{FIG_WEIGHT_MODEL} (b), the \gls{bb} is modelled as a single broad distribution, capturing the probability of having spurious detections.

Bringing it all together, the \emph{per-frame} weight is 
\begin{equation}
\MMseTrackWeight_{i, j}^{[f]} = \log \Prob{BB_i^{[f]} | p_j^{[f]},c_j^{[f]}} ,
\end{equation}
with:
\begin{equation}
\Prob{BB_i | p_j, c_j} = 
\begin{cases}
	\sum_{v_j} \Prob{BB_i | p_j, v_j} \Prob{v_j | p_j, c_j} & \text{ if } j < J+1 \\ 
\Prob{BB_i | \OObject_J} & \text{ otherwise } 
\end{cases}, \label{EQ_PER_FRAME_WEIGHT}
\end{equation}
where we have dropped the frame index for clarity.
The complete tracklet-object weight $\MMseTrackWeight_{i, j}$ is then the sum of the log-probabilities over the tracklet's visible lifetime.

\subsection{Relation to other methods}
A number of works (see \eg \cite{MTH_019, VL_067, VL_071, VL_072}) have used \gls{ilp} to address the tracking problem, however we reiterate that our task goes beyond tracking to data association with the external information.
Indeed, in all of the above cited approaches, the emphasis is on joining the successive detections into contiguous tracklets, while our focus is on the subsequent step of identifying the tracklets by way of the \gls{rfid} position.

Solving identification problems similar to our is typically achieved through hypothesis filtering, \eg \gls{jpda} \cite{MTH_019}.
While we could have posed our problem into this framework (which would be a contribution in its own right, with the inclusion of the position-affinity model in \sect \ref{SS_ASSIGNMENT_WEIGHTING}), exact inference for such a technique increases exponentially with time.
While there are approximation algorithms (\eg \cite{VL_053}), our ILP formulation:
\begin{enumerate*}[label=(\alph*)]
\item \label{IT_JPDA_NO_TIME} does not explicitly model temporal continuity (by delegating this to the tracker),
\item \label{IT_JPDA_INDEPENDENT} imposes the restriction that the (per-frame) weight of an animal-tracklet pair is a function only of the pair (\ie independent of the other assignments),
\item \label{IT_JPDA_MAP} estimates a \emph{\gls{map}} tracklet weight rather than maintaining a global distribution over all possible tracklets, \textsl{but then}
\item \label{IT_GLOBAL} optimises for a globally consistent solution rather than running online frame-by-frame filtering.
\end{enumerate*}
The combination of \ref{IT_JPDA_NO_TIME}, \ref{IT_JPDA_INDEPENDENT} and \ref{IT_JPDA_MAP} greatly reduce the complexity of the optimisation problem compared to \gls{jpda}, allowing us to optimise an offline global assignment \ref{IT_GLOBAL} without the need for simplifications/approximations.

%-------------------------------------------------------------------------%

\section{Implementation details}
\label{S_IMPLEMENTATION}

\subsection{Detector}
While the detector is not the core focus of our contribution it is an important first component in our identification pipeline.

We employ the widely-used FCOS object detector \cite{VL_060} with a ResNet-50 \cite{NA_041} backbone, although our method is agnostic to the type of detector as long as it outputs \glspl{bb}.
We replace the classifier layer of the pre-trained network with three categories (\ie,~\textit{mouse, tunnel and background}) and fine-tune the model on our data (see \supp \sect C.1).
In the final configuration, we accept all detections with confidence above 0.4, up to a maximum of 5 per image.
We further discard \glspl{bb} which fall inside the hopper area by thresholding on a maximum ratio of overlap (with hopper) to \gls{bb} area of 0.4.
This setup achieves a recall (at \gls{iou}=0.5) of 0.90 on the held-out test-set (average precision is 0.87, CoCo \gls{mnap}=0.60).
Given our focus on the rest of the identification pipeline, we leave further optimisation over architectures to future work.

\subsection{Tracker}
We use the SORT tracker of Bewley \etal \cite{VL_051} which maintains a `latent' state of the tracklet (the centroid $(x,y)$ of the \gls{bb}, its area and the aspect ratio, together with the velocities for all but the aspect ratio) and assigns new detections in successive frames based on \gls{iou}.
At each frame, the state of each active tracklet is updated according to the Kalman predictive model, and new detections matched to existing tracklets by the Hungarian algorithm \cite{VL_051} on the \gls{iou} (subject to a cutoff).

Our plug'n'play approach would allow us to use other advanced trackers (\eg \cite{VL_052, VL_063, VL_071, VL_064}), but we chose this tracker for its simplicity.
Since most of the above methods leverage visual appearance information, and all mice are visually similar, it is likely that any marginal improvements would be outweighed by a bigger impact on scalability.
In addition, previous research (\eg \cite{DS_018}) has shown that simple methods often outperform more complex ones, but we leave experimentation with trackers as possible future research.

We do, however, make some changes to the architecture, in light of the fact that we do not need online tracking, and hence can optimise for retrospective filtering.
The first is that tracklets emit \glspl{bb} from the start of their lifetime, and we do not wait for them to be active for a number of frames: this is because, we can always filter our spurious detections at the end.
Secondly, when filtering out short tracklets, we do this based on a minimum number of contiguous (rather than total) frames --- again, we can do this because we have access to the entire lifetime of each tracklet.
We also opt to emit the original detection as the \gls{bb} rather than a smoothed trajectory, since we believe that the current model assuming a fixed aspect ratio might be too restrictive for our use-case --- indeed, we plan to experiment with different models for the Kalman state in the future.
Finally, we take the architectural decision to kill tracklets as soon as they fail to capture a detection in a frame: our motivation is that we prefer to have broken tracklets which we can join later using the identification rather than increasing the probability of a tracklet switching identity.

To optimise the two main parameters of the tracker --- the \gls{iou} threshold for assigning new detections to existing tracklets, and the minimum lifetime of a tracklet to be considered viable --- we used the combined training/validation set, and ran our \gls{ilp} object identification end-to-end over a grid of parameters.
The optimal results were obtained using a length cutoff of 2 frames and \gls{iou} threshold of 0.8 (see \supp \sect C.2 for details).

\subsection{Weight model}

We represent \glspl{bb} by the 4-vector containing the centroid $[x, y]$ and size $[width, height]$ of the axis-aligned \gls{bb}.
A hidden \gls{bb} is denoted by the zero-vector and is handled explicitly in logic.
The visibility, $v_j$ can be one of clear, truncated or hidden as already discussed, while the position, $p_j$ is an integer representing the one-of-18 antennas in which the mouse is picked up.
For $c_j$ we consider the \gls{rfid} pickups of the other two mice, but we must represent them in an (identity) permutation-invariant way: \ie the representation must be independent of the identify of the animal we are predicting for (all else being equal).
To this end, we define a nine-point neighbourhood around the antenna $p_j$ as in Fig.\ \ref{FIG_WEIGHT_MODEL}(c), where $p_j$ would fall in cell 4.
Note that where the neighbourhood extends beyond the range of the antenna base-plate, these cells are simply ignored. $c_j$ is then simply the count of animals picked up in each cell (0, 1 or 2).

Under the outlier model (Fig.\ \ref{FIG_WEIGHT_MODEL} (b)), the \gls{bb} centroids, $[x, y]$, are drawn from a very wide Gaussian over the size of the image: the size, $[w, h]$ is Gaussian with its mean and covariance fit on the training/validation annotations.
For the non-outlier case (Fig.\ \ref{FIG_WEIGHT_MODEL}(a)) we model the distribution $p(v_j|p_j,c_j)$ using a \gls{rf}: this proved to be the best model in terms of validation log-likelihood (this metric is preferable for a calibrated distribution).
The mean $\mu_{pv}$ of the multivariate Gaussian for the \gls{bb} parameters at position $p$ with visibility $v$ in Eq.\ \eqref{EQ_BB_PROB} is defined as follows.
The centroid components $[x, y]$ are the same irrespective of the visibility, and governed by a homography mapping between the antenna positions on the ground-plane and the annotated \glspl{bb} centroids in the image.
The width and height, $[w, h]$ are estimated independently for each row in the antenna arrangement (see Fig.\ \ref{FIG_DATASET_EXAMPLE} (b)) and visibility (clear/truncated).
This models occlusion and perspective projection but allows us to pool samples across pickups for statistical strength.
The covariance matrix $\Sigma_{pv}$ is estimated solely on a per-row basis, and is independent of the visibility. Further details on fine-tuning the parameters appear in our \supp (\sect C.3).

\subsection{Solver}
Unfortunately, our linear program is not Totally Unimodular \cite{MTH_017} due to the constraints in Eqs.\ (\ref{EQ_TRK_LPC_1MSE} -- \ref{EQ_TRK_INVALID}), and hence, we have to explicitly enforce integrality.
Consequently, solving the \gls{ilp} is in general NP-Hard, but we have found that due to our temporal abstraction, using a branch and cut approach efficiently finds the optimal solution on the size of our data.
We model our problem through the Python MIP package \cite{MISC_057}, using the COIN-OR Branch and Cut \cite{MISC_058} paired with the COIN-OR Linear Programming \cite{MISC_059} solvers.
Running on a conventional desktop (Intel Xeon E3-1245 @3.5GHz with 32Gb RAM), the solver required on the order of a minute on average for each 30-minute segment.

\section{Experiments}
\label{S_EXPERIMENTS}

\subsection{Dataset}
\label{SS_EXP_DATASET}
\paragraph{Source:}
We apply and evaluate our method on a mouse dataset, \cite{CBD_026}, provided by \gls{mrc}.
All the mice in the study are one-year old males of the C57BL/6Ntac strain.
The mice are housed as groups of three in a cage, and are continuously recorded for a period of 3 to 4 days at a time, with a 12-hour lights-on, 12-hour lights-off cycle. We have access to 15 distinct cages.
The recordings are split into 30-minute periods, which form our unit of processing (segments). 
Since we are interested in their crepuscular rhythms, we selected the five segments straddling either side of the lights-on/off transitions at 07:00 and 19:00 every day.
This gives us about 30 segments per cage.

The data --- \gls{ir} side-view video and \gls{rfid} position pickup --- is captured using one of four \gls{hca} rigs from \acta: a prototypical setup is shown in Fig.\ \ref{FIG_HCA_RIG} (a).
Video is recorded at 25 \gls{fps}, while the antenna baseplate ($3\times 6$ cells, see Fig.\ \ref{FIG_HCA_RIG} (b)) is scanned at a rate of about 2Hz and upsampled to the frame rate.
The latter, however, is noisy due to mouse huddling and climbing.
The video also suffers from non-uniform lighting, occasional glare/blurring and clutter, particularly due to a movable tunnel and bedding.
Most crucially, however, being single strain (and consequently of the same colour), and with no external markings, the mice are visually indistinguishable.

\paragraph{Pre-processing:}
A calibration routine was used to map all detections into the same frame of reference using a similarity transform.
In order to maintain axis-aligned \glspl{bb}, we transform the four corners of the \gls{bb} but then define the transformed \gls{bb} to be the one which intersects the midpoints of each transformed edge (we can do this because the rotation component is very minimal, $<2^\circ$).
A minimal number of segments which exhibited anomalous readings (\eg spurious/missing \gls{rfid} traces) that could not be resolved were discarded altogether  (see \supp \sect A.1).

\paragraph{Ground-truthing:}
Using the VIA tool \cite{MISC_054}, we annotate video frames at four-second intervals for three-minute `snippets' at the beginning, middle and end of each segment, as well as on the minute throughout the segment.
This gives a good spread of samples with enough temporal continuity to evaluate our models.
The annotations consist of an axis-aligned \gls{bb} for each visible mouse, labelling the identity as Red/Green/Blue and the level of occlusion as clear \vs truncated: hidden mice are by definition not annotated.
A \emph{Difficult} flag is set when it is hard to make out the mouse even for a human observer.
Where the identity of any mouse cannot be ascertained with certainty, this is noted in the schema and the frame discarded in the final evaluation.
Refer to \supp \sect A.2 for further details.

\paragraph{Data splits:}
Our 15 cages were stratified into training (7), validation (3) and testing (5). This guarantees unbiased generalisation estimates, and shows that our method can work on entirely novel data (new cages/identities) without the need for additional annotations.
We annotated a random subset of 7 segments (3.5 Hrs) for training, 6 (3 Hrs) for validation and 10 (5 Hrs) for testing (2 per-cage).
This yielded a total of 740 (training), 558 (validation) and 766 (testing) frames in which the mice could be unambiguously identified.
We used all frames within the training/validation sets for optimising the weight model, but the rest of the architecture is trained and evaluated on three-minute snippets only, using 498, 398 and 699 frames respectively (for testing we remove snippets in which 50\% or more of the frames show immobile mice or tentative/huddles due to severe occlusion.)

\subsection{Comparison Methods}
We compare the performance of our architecture to two simpler baseline models and off-the-shelf trackers. For overall statistics, we also report the maximum possible achievable performance given the detections.

The first of the baselines, \emph{Static (C)}, is a per-frame assignment (Hungarian algorithm) based on Euclidean distance between the centroid of the \glspl{bb} and the projected \gls{rfid} tag.
The second model, \emph{Static (P)}, uses the probabilistic weights of \sect \ref{SS_ASSIGNMENT_WEIGHTING}, but is assigned on a per-frame basis.

Our problem is quite unique for the reasons enumerated in \sect \ref{S_RELATED_WORK}, and hence, off-the-shelf solutions do not typically apply.
Most systems are either designed for singly-housed mice \cite{VL_058, VL_057} or leverage different fur-colours/external markings \cite{VL_056, MS_001} to identify the mice.
We ran some experiments with \emph{idtracker.ai} \cite{VL_055} but it yielded no usable tracks --- the poor lighting of the cage, coupled with the side-view camera interfered with the background-subtraction method employed.
Bourached \etal \cite{VL_045} showed that the widely-used DeepLabCut \cite{VL_059} framework does not work on data from the Harwell lab, possibly because of the side-view causing extreme occlusion of key body parts.
The newer version of DeepLabCut \cite{VL_069} supports multi-animal tracking, but for identification it requires supervision with annotated samples (full keypoints) for each new cage.
Such a level of supervision would be highly onerous and impractical.
In contrast our method can be applied to new cages without retraining, as long as the visual configuration is similar (and indeed, we test it on cages for which we have not trained on).
As a proof of concept, we explored annotating body-parts for one of the cages, but found it impossible to identify reproducible body-parts due to the frequent occlusions.
For this reason, we cannot report any results for any of the above methods.

It should be emphasized that there is a misalignment of goals when it comes to comparing our method with \acrlong{sota} \gls{mot} trackers --- no matter the quality of the tracker, there is still the need for an identification stage, and hence, the comparison degenerates to optimising over tracker architectures (which is outside the scope of our research).
For example, while DeepLabCut does support multi-animal scenarios, it requires a consistent identification across videos (possibly using visual markings) to support identifying individuals: otherwise, it generates simple tracklets.

However, by way of exploration, we can envision having a human-in-the-loop that seeds the track (if the tracker supports this) and provides the identifying information.
To simulate this, we took each three-minute snippet, extracted the ground-truth annotation at the middle frame, and initialised a SiamMask \cite{VL_028} tracker per-mouse, which we ran forwards and backwards to either end of the snippet.
We used this architecture as it is designed to be a general purpose tracker with no need for pre-training on new data, and hence was readily usable with the annotations we already had.

\subsection{Evaluation}
The standard \gls{mot} metrics (\gls{motp} and \gls{mota}) \cite{VL_013} do not sufficiently capture the needs of our problem, in that 
\begin{enumerate*}[label=(\alph*)]
\item we have a fixed set of objects we need to track (making the problem better defined), \textsl{but}
\item we also care about the absolute identity (\ie permutations are not equivalent).
\end{enumerate*}
While there is a similarity with object detection \cite{DS_016, DS_015} (using identity for object class), we know a priori that we cannot have more than one detection per `identity' in each frame, and thus we modify existing metrics to suit our constraints.

\paragraph{Overall performance:}
For each (annotated) frame $f$ and animal $j$, we define a groundtruth \gls{bb}, $GT_j^{[f]}$ --- this is null when the animal is hidden.
The identifier itself outputs a single hypothesis $\smash{BB_j^{[f]}}$ for animal $j$, which can also be null.
Subsequently, the `Overall Accuracy', $A_{O}$, is the ratio of ground-truths that are correctly identified: \ie where (a) the object is visible and the \gls{iou} between it and the identified \gls{bb} is above a threshold, or (b) when the object is hidden and the identifier outputs null.
Mathematically, this is:
\begin{equation}
    A_{O} = \frac{1}{FJ}\sum_{f=1}^{F}\sum_{j=1}^{J} \Delta \left( BB_{j}^{[f]}, GT_{j}^{[f]}\right), \label{EQ_OVERALL_ACCURACY}
\end{equation}
where
\begin{equation}
    \Delta \left( B, G\right) \equiv \begin{cases}
        1 & \text{if } B = \emptyset, G = \emptyset, \\
        1 & \text{if } IoU\left(B, G\right) > \phi, \\
        0 & \text{otherwise} .
    \end{cases}
\end{equation}
In the above, we assume that \gls{iou} is 0 if any of the \glspl{bb} is null: we use an \gls{iou} threshold, $\phi$ of 0.5 for normal objects, and 0.3 for annotations marked as \emph{Difficult}.

A separate score, the `Overall \gls{iou}', captures the average overlap between correct assignments, and is defined only for objects which are visible:
\begin{equation}
    \text{\gls{iou}}_O = \frac{1}{\sum_{f=1}^{F}J^{[f]}}\sum_{f=1}^{F}\sum_{j=1}^{J} IoU\left(BB_j^{[f]}, GT_j^{[f]}\right) ,
\end{equation}
where $J^{[f]}$ represents the number of ground-truth visible mice in frame $f$: again, we assume that the \gls{iou} is 0 if any \gls{bb} is null.

We also report some `binary' metrics.
The \gls{fnr} is given by the average number of times a visible object is not assigned a \gls{bb} by the identifier:
\begin{equation}
    \text{\gls{fnr}}_{O} = \frac{1}{\sum_F J^{[f]}} \sum_{f=1}^{F}\sum_{j=1}^{J}\left(BB_j^{[f]} = \emptyset \land GT_j^{[f]} \neq \emptyset \right) ,
\end{equation}

where $\land$ is the \emph{logical and}.
Note that this metric does not consider whether the predicted \gls{bb} is correct: this is handled by the `Uncovered Rate' which augments the \gls{fnr}$_)$ by measuring the average number of times a visible object is assigned a \gls{bb} that does not `cover' it:
\begin{equation}
\text{U}_{O} = \frac{1}{\sum_F J^{[f]}} \sum\limits_{f=1}^{F}\sum\limits_{j=1}^{J}\left(BB_j^{[f]} \neq \emptyset \land GT_j^{[f]} \neq \emptyset \land IoU\left(BB_j^{[f]}, GT_j^{[f]}\right) < \phi \right) . \label{EQ_EXP_UNCOVERED}
\end{equation}
Finally, the \gls{fpr} counts the (average) number of times a hidden object is wrongly assigned a \gls{bb}:
\begin{equation}
        \text{\gls{fpr}}_{O} = \frac{1}{\sum_{f=1}^{F} \overline{J^{[f]}}} \sum_{f=1}^{F}\sum_{j=1}^{J} \left(BB_j^{[f]} \neq \emptyset \land GT_j^{[f]} = \emptyset \right) , \label{EQ_OVERALL_FNR}
\end{equation}
where $\overline{J^f}$ represents the number of animals hidden at frame $f$.

\paragraph{Performance conditioned on detections:}
The above metrics conflate the localisation of the \gls{bb} (\ie performance of the detector) with the identification error, which is what we actually seek to optimise.
We thus define additional metrics conditioned on an oracle assignment of detections to ground-truth.
Under the oracle, each candidate detection $i$ is assigned to the `closest' (by \gls{iou}) ground-truth annotation $j$ using the Hungarian algorithm  with a threshold of 0.5: this is relaxed to 0.3 for detections marked as \emph{Difficult}.
The identity of the detection $O_{i}^{[f]}$ is then that of the assigned ground-truth or, null if not assigned.
Given the identity $ID_{i}^{[f]}$ assigned by our identifier to \gls{bb} $i$, we define the Accuracy given Detections as the fraction of detections which are given the correct (same as oracle) assignment by the identifier (including null when required):
\begin{equation}
    A_{GD} = \frac{1}{\sum_{f=1}^F I^{[f]}}\sum_{f=1}^{F}\sum_{i=1}^{I^{[f]}} \delta\left(ID_{i}^{[f]}, O_{i}^{[f]}\right),
\end{equation}
where $I^{[f]}$ is the number of detections in Frame $f$, and $\delta$ is the Kronecker delta (\ie $\delta(a,b) = 1$ iff $a = b$ and 0 otherwise).

We also report the rate of Mis-Identifications (\ie a wrong identity is assigned): these are \glspl{bb} which are given an identity by the oracle and a different identity by the Identifier.
\begin{equation}
\text{MisID}_{GD} = \frac{\sum_{f=1}^{F}\sum_{i=1}^{I^{[f]}} \left(ID_{i}^{[f]} \neq \emptyset \land O_{i}^{[f]} \neq \emptyset \land ID_{i}^{[f]} \notin O_{i}^{[f]} \right)}{\sum_{f=1}^{F}\sum_{i=1}^{I^{[f]}} \left(O_{i}^{[f]} \neq \emptyset \right)} \label{EQ_EXP_MISID}
\end{equation}
For the purpose of this metric, null assignments are not counted as erroneous, and we normalise the count by the number of non-null oracle assignments.

Always relative to the oracle, we also quote the \gls{fpr} and \gls{fnr}.
The \gls{fpr} is now defined as the fraction of \glspl{bb} with a null oracle assignment that are assigned an identity:
\begin{equation}
   \text{\gls{fpr}}_{GD} = \frac{\sum_{f=1}^{F}\sum_{i=1}^{I^{[f]}} \left(ID_{i}^{[f]} \neq \emptyset \land O_{i}^{[f]} = \emptyset \right)}{\sum_{f=1}^{F}\sum_{i=1}^{I^{[f]}} \left(O_{i}^{[f]} = \emptyset \right)} ,
\end{equation}
and conversely, the ratio of \glspl{bb} with a non-null oracle assignment that are not identified define the \gls{fnr}:
\begin{equation}
	\text{\gls{fnr}}_{GD} = \frac{\sum_{f=1}^{F}\sum_{i=1}^{I^{[f]}} \left(ID_{i}^{[f]} = \emptyset \land O_{i}^{[f]} \neq \emptyset \right)}{\sum_{f=1}^{F}\sum_{i=1}^{I^{[f]}} \left(O_{i}^{[f]} \neq \emptyset \right)} . \label{EQ_FNR_GD}
\end{equation}
In fact, if we consider raw counts (rather than ratios), the sum of Mis-Identifications, \gls{fpr} and \gls{fnr} constitute all the errors the system makes.

\paragraph{Evaluation data-size:}
Table \ref{TAB_DATA_SIZE} summarises the size of each of the normalisers (within each dataset) that are used to compute the above metrics: the symbols used are the same as those used in the normalisers of Eqs.\ \ref{EQ_OVERALL_ACCURACY} -- \ref{EQ_FNR_GD}.

\begin{table}[]
    \centering
    \begin{tabular}{lcccccc}
    	% Generated using `Evaluate_Identifiers.ipynb`:
    	%    -- Overall -- Tune: S3.3.1, Test: S5.2.1
    	%    --Detection-- Tune: S3.4.1, Test: S5.2.1
        \toprule
        {} & \multicolumn{3}{c}{Overall} & \multicolumn{3}{c}{Given detections} \\
             \cmidrule(lr){2-4}  		   \cmidrule(lr){5-7}
        {} & $F J$ & $\sum_F J^{[f]}$ & $\sum_F\overline{J^{[f]}}$  & $\sum_{F} I^{[f]}$ & $\sum_{FI^{[f]}} O_i^{[f]} \neq \emptyset$ & $\sum_{FI^{[f]}} O_i^{[f]} = \emptyset$ \\
        \midrule
        Tune &    2688 &    2541 &    147 &     2967 &       2235 & 732 \\
        Test        &    2097 &    2012 &     85 &   2504 &     1836 & 668 \\
        \bottomrule
    \end{tabular}
    \caption{Number of samples for evaluating the models according to the Overall and Given Detections metrics in the Tuning (Train + Validation) and Test sets respectively.}
    \label{TAB_DATA_SIZE}
\end{table}

\subsection{Results}
We comment on the quantitative performance of our approach.
We also make available a sample of video-clips showing our framework in action at \repo: we comment on these in the \supp (see \sect D).

\paragraph{Performance on the test set:}
Table \ref{TAB_RESULTS_TEST} shows the results of our model and comparative architectures evaluated on the held-out test-set (the number of samples in each case appear in Table \ref{TAB_DATA_SIZE}), where our scheme clearly outperforms all other methods on all metrics apart from \gls{fnr}$_{O}$.
Table \ref{TAB_COUNT_RESULTS} on the other hand shows the raw counts of the same metrics as Table \ref{TAB_RESULTS_TEST}.
Looking first at the left-side of the table, the overall accuracy is quite high at 77\%.
To get a feel of what this means consider that we are limited by the recall of the detector.
In fact, the oracle, which represents an upper bound (given the detector), scores 94\%.
The \gls{fpr} seems high, but in reality, the number of hidden samples (see Table \ref{TAB_COUNT_RESULTS}) is only 85 (4\%).

\begin{table}[!ht]
    \centering
    \setlength\tabcolsep{1.3mm}
	\begin{tabular}{lccccccccc}
	% Generated using `Evaluate_Identifiers.ipynb`
	%    - All Models: S5.2.2
	%    - Oracle: S1
    \toprule
        {} & \multicolumn{5}{c}{Overall} & \multicolumn{4}{c}{Given Detections} \\
    \cmidrule(r){2-6}\cmidrule(l){7-10}
        {Model} & $A_O\uparrow$ & $\text{\gls{iou}}_O\uparrow$  & $U_O\downarrow$ & FNR $\downarrow$ &   FPR $\downarrow$ &         $A_{GD}\uparrow$ & MisID $\downarrow$ &   FNR $\downarrow$ &   FPR $\downarrow$ \\
        \midrule
        Static (C)    &    0.659 & 0.626 & 0.286 & 0.036 & 0.788 &    0.623 &  0.178 & 0.122 & 0.590 \\
        Static (P)    &    0.716 & 0.666 & 0.219 & 0.045 & 0.741 &    0.694 &  0.138 & 0.100 & 0.496 \\
        Ours          &   \textbf{0.767} & \textbf{0.694} & \textbf{0.145} & 0.070 & \textbf{0.659} &            \textbf{0.791} &    \textbf{0.104} & \textbf{0.066} & \textbf{0.317} \\
SiamMask     & 0.637 &   0.565 & 0.336 & \textbf{0.000} & 1.000 &     --- &      --- &   --- &   --- \\
        \midrule
 Oracle		 & 0.916 & 0.770 & 0.000 & 0.087 & 0.000 & --- &      --- &   --- &   --- \\
    \bottomrule
	\end{tabular}
    \caption{Comparative Results on the Test-Set for the Overall Accuracy, and
Accuracy Given Detections. \emph{Static (C)} is a baseline, \emph{Static (P)} uses our weight model on a per-frame basis and the full temporal identifier is labelled \emph{Ours}. We also report the overall performance of the SiamMask architecture as well as the oracle assignment as a theoretical upper-bound on performance.}
    \label{TAB_RESULTS_TEST}
\end{table}

The \gls{fnr}$_O$ is the only overall metric in which our method suffers, mostly because it prefers not to assign a \gls{bb} rather than assign the wrong one: this is evidenced instead by the lower rate of Uncovered.
Note that in moving from the baseline through to our final method, the performance strictly improves each time for all other metrics.
Indeed, a big jump is achieved simply by the use of the weight model.
On the other hand, the performance of the SiamMask \cite{VL_028} architecture is comparable to the Static (C) baseline.
It should be noted that such a method would require a level of ground-truth annotation per-video to jump-start the process (\ie a human-in-the-loop), but then could have been a viable baseline given that it requires no fine-tuning, while our detector needed to be trained using annotations.
However, its performance is inadequate mostly because (a) it is not tailored to tracking mice and (b) because it cannot reason about occlusion, leading to high-levels of false positives.

The contrast between methods is starker when it comes to the analysis given detections (the SiamMask cannot be compared because it does not utilise detections).
This can be explained because in the former, the performance of the detector acts as an equaliser between alternatives, while $A_{GD}$ teases out the capability of the identification system from the quality of the detector.
Here the \gls{fpr}$_{GD}$ is lower, although from the point of view of detections, there are more of them that should be null (1563 or 45\%), and hence, the ability of the identification system to reject spurious detections is illustrated.
Note that the number of background \glspl{bb}, (impacting the \gls{fpr}$_{GD}$) is higher.
We also investigated outlier models which were fit explicitly to the outliers in our data but validation-set evaluation yielded a drop in performance.

\begin{table}[]
    \centering
    \setlength\tabcolsep{1.5mm}
    % Generated using `Evaluate_Identifiers.ipynb` (S5.2.2)
    \begin{tabular}{lcccccccc}
        \toprule
        {}       & \multicolumn{4}{c}{Overall} & \multicolumn{4}{c}{Given Detections} \\
        \cmidrule(r){2-5}\cmidrule(l){6-9}
        {Model} & $A_O\uparrow$ & $U_O\downarrow$ & FNR $\downarrow$ &   FPR $\downarrow$ &         $A_{GD}\uparrow$ & MisID $\downarrow$ &   FNR $\downarrow$ &   FPR $\downarrow$ \\
        \midrule
        Static (C) & 1382 &  575 & 73  & 67 &  1559 &  327 &  224 &  394 \\
        Static (P) & 1502 &  441 & 91  & 63 &  1737 &  253 &  183 &  331 \\
        Ours       & \textbf{1608} &  \textbf{292} & 141 & \textbf{56} &  \textbf{1980} &  \textbf{191} &  \textbf{121} &  \textbf{212} \\
        SiamMask   & 1335 &  677 & \textbf{0} & 85 &   --- &  --- &  --- &  --- \\
        \midrule
        Normaliser & 2097 & 2012 & 2012 & 85 &  2504 & 1836 & 1836 & 668 \\
        \bottomrule
    \end{tabular}
    \caption{Counts of Performance metrics on the Test-Set (counterpart to Table 1). \gls{iou} has no meaning as a count. The last row shows the normaliser which would be used in computing rates.}
    \label{TAB_COUNT_RESULTS}
\end{table}

\begin{figure}[!h]
    \centering
    \includegraphics[width=0.92\textwidth]{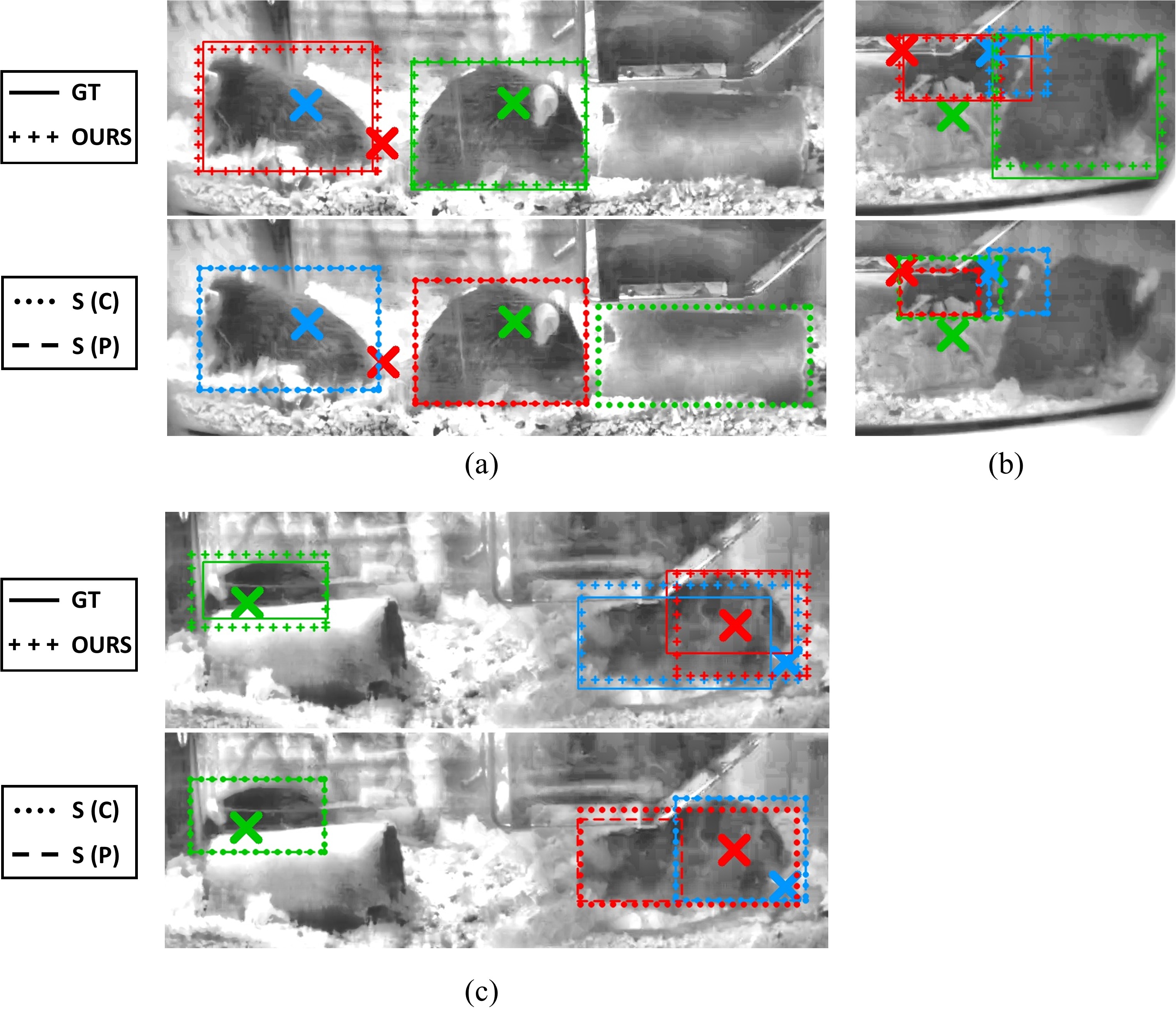}
    \caption{
    Examples of successful identifications for OUR method (cropped to appropriate regions and brightened with CLAHE for clarity). 
    Mouse pickups are marked by crosses of the appropriate colour.
    In each panel, (a)--(c), the ground-truth (\textbf{GT}) and identifications according to our model (\textbf{OURS}) appear in the top row, while the bottom part shows the identifications due to the centroid-distance (\textbf{S (C)}) and static probabilistic (\textbf{S (P)}) models.
    }
    \label{FIG_OUTSTANDING}
\end{figure}

\paragraph{Example of Successful Identifications:}
Fig.\ \ref{FIG_OUTSTANDING} shows a visualisation of the identification using our method (top) and the baseline/ablation models (below) for three sample frames.
Panel (a) shows an occlusion scenario: the Blue mouse is entirely hidden by the Red one.
This throws off the Static (C) method, to the extent that it classifies all mice incorrectly and picks up the tunnel as the Green mouse.
The observation model in the Static (P) ablation is able to reject this spurious tunnel detection, but is confused by the other two detections, with the Red mouse marked Blue and the Green mouse identified as Red.
Our method is able to use the temporal context to reason about this occlusion and correctly identifies Red/Green and rejects Blue as Hidden.
A similar phenomenon is manifested in panel (c), where the temporal context present in our method (OURS) helps to correctly make out Red/Blue but the baseline/ablation get them mixed up: Static (P) particularly, uses the wrong \gls{bb} altogether (which covers half a mouse).

Moving on to panel (b), this illustrates a particularly challenging scenario due to the three mice being in very close proximity under the hopper.
In this case, annotation was only possible after repeated observations and following through of the \gls{rfid} traces: indeed, while Green and Red are somewhat visible, Blue is almost fully occluded and is annotated as \emph{Difficult}.
Most significantly, however, the \gls{rfid} pickups are shifted although in the correct relative arrangement.
The baseline/ablation methods completely miss out the Green mouse due to this, and while Blue is assigned a correct \gls{bb}, Red is confused as Green, with the Red \gls{bb} covering a spurious detection.
Despite this, our method correctly identifies all three mice, even if the \glspl{bb} for Red and Blue do not perfectly cover the extent of the respective mice.

\begin{figure}[!h]
    \centering
    \includegraphics[width=0.98\textwidth]{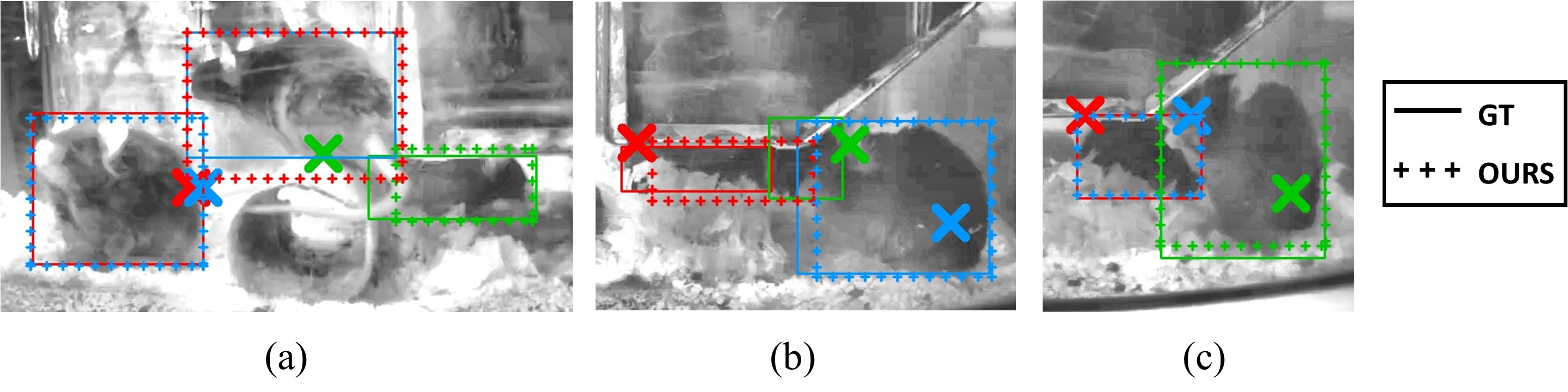}
    \caption{
    Visualisation of failure cases illustrating (a) switched identification, (b) false negative and (c) false positive cases.
    The markups follows the arrangement in Fig.\ \ref{FIG_OUTSTANDING}.
    }
    \label{FIG_PATHOLOGICAL}
\end{figure}

\paragraph{Sample Failure Cases}
Despite outperforming the competition, our model is not perfect.
We show some such failure cases in Fig.\ \ref{FIG_PATHOLOGICAL}.
Focusing first on panel (a), the switched identification between the Red and Blue mice happens due to a lag in the \gls{rfid} pickup.
In this case, Blue has climbed on the tunnel, but this is too high to be picked up by the baseplate and hence its position remains at its original location, occupying the same antenna space as Red.
When this happens, symmetry is broken solely by the temporal tracking, but it appears this failed in this case.

Panel (b) shows a challenging huddling scenario, with the mice in close proximity: indeed, the Green mouse is barely visible, as it is hid by the Red mouse and the hopper.
This produces a false-negative: our method is unable to pick up the Green mouse, classifying it as Hidden, altought it correctly identifies Red and Blue.
A similar scenario appears in panel (c), although this time, the method also causes a false positive.
In (c), the Red mouse is completely hidden behind the Blue/Green mice.
Our method picks up green correctly, but assigns Red's \gls{bb} to Blue, effectively incurring two errors: a false positive (Blue should be occluded) and a false negative (Red is not picked up).

The last example also illustrates the interdependent nature of the errors, a side-effect of the constraint that there are at most one of each `colour' of mice.
This justifies our use of specific metrics to mitigate this overlap: note how our definition of Uncovered (Eq.\ \ref{EQ_EXP_UNCOVERED}) and Misidentification (Eq.\ \ref{EQ_EXP_MISID}) explicitly discounts such instances as they would have already been captured by false negatives/positives.

%The scenario in the next frame is particularly challenging due to the mice huddling in close proximity.
%Although not perfect, the Static (P) assignment is able to reason about the relative arrangement of the Red/Blue mice, including that Blue occludes Green: however, the temporal aspect throws off our model.

%\modif{
%It is interesting to understand what tends to hurt performance in our method.
%In the first panel on the right of the legend in Fig.\ \ref{FIG_PATHOLOGICAL}, the Green mouse is barely visible (marked as `Difficult').
%However, due to the absolute distances, the centroid approach picks up the correct identities.
%On the other hand, because it is such an extreme case, the probabilistic weight confuses Blue and Green, although our full temporal method at least disregards the small \gls{bb} rather than assigning the wrong identity.
%On the flip-side, the weight model is able to leverage the statistics it inferred from the data to improve identification, as in the rightmost panel.
%}
%Note how due to the Gaussian model on \gls{bb} centroid and size, it is able to correctly pick out the Red mouse \modif{and disregard the tunnel detection}.
%This is also what allows it to reason about the occluded Blue mouse huddling with Green, despite the \gls{rfid} seemingly showing the contrary.

%-------------------------------------------------------------------------%

\section{Conclusions}
\label{S_CONCLUSION}

We have proposed an \gls{ilp} formulation for identifying visually similar animals in a crowded cage using weak localisation information.
Using our novel probabilistic model of object detections, coupled with the tracker-based combinatorial identification, we are able to correctly identify the mice 77\% of the time, even under very challenging conditions.
Our approach is also applicable to other scenarios which require fusing external sources of identification to video data.
We are currently looking towards extending the weight model to take into account the orientation of the animals as this allows us to reason about the relative position of the \gls{bb} to the \gls{rfid} pickup, as well as modelling the tunnel which can occlude the mice.

\subsection*{Supplementary Material:}
We make available the curated version of the dataset and code at \repo, with instructions on how to use it.
We provide two versions of the data: a detections-only subset to train and evaluate the detection component, as well as an identification subset.
In both cases, we provide ground-truth annotations.
For detections we provide the pre-extracted frames.
For identification, in addition to the raw video and position pickups, we also released the \glspl{bb} as generated by our trained FCOS detector, allowing researchers to evaluate the methods without the need to re-train a detector.
Further details about the repository are provided in Appendix \ref{APP_SUPP}.

% Back Matter
\subsection*{Acknowledgements}
We thank our collaborators at the \acrlong{mrc}, especially Dr Sara Wells and Dr Pat Nolan for providing and explaining the data set.

\subsection*{Funding}
MC's work was supported by the EPSRC Centre for Doctoral Training in Data Science, funded by the UK Engineering and Physical Sciences Research Council (grant EP/L016427/1) and the University of Edinburgh.
LZ was supported by NSFC (No. 62106050).
RSB was supported by funding to \gls{mrc} from the Medical Research Council UK (grant A410).
Funding was also provided by the EPSRC Programme Grant Seebibyte EP/M013774/1.

\subsection*{Ethical Approval}
Data collection for all animal studies and procedures used in the mouse dataset were carried out at \gls{mrc}, in accordance with the Animals (Scientific Procedures) Act 1986, UK, Amendment Regulations 2012 (SI 4 2012/3039) as indicated in \cite{CBD_026}.

%------------------------------------------------------%

\clearpage
\begin{center}
\Huge\textsc{Appendices}
\end{center}
\appendix
\counterwithin{table}{section}
\counterwithin{figure}{section}
\counterwithin{equation}{section}
\section{Dataset}
\label{APP_DATASET}
In this section we elaborate on the Dataset used in our work, including the annotation schema used.

\subsection{Data cleaning and calibration}
\label{SS_DATA_CLEANING}
We removed one of the cages as it had a non-standard setup.
The \gls{rfid} metadata was then filtered through consistency logic to ensure that the same mice were visible in all recordings from the same cage and that there was no missing data.
In some cases, missing data due to huddling mice (which causes the \gls{rfid} to be jumbled) could be filled in by looking at neighbouring periods to see if the mice moved in between and if not the same position filled in: otherwise, the segment was discarded.
Segments in which there was a significant timing mismatch between recording time and \gls{rfid} traces (indicating a possible invalid trace) were also discarded.

\begin{figure}[!ht]
    \centering
    \includegraphics[width=0.95\textwidth]{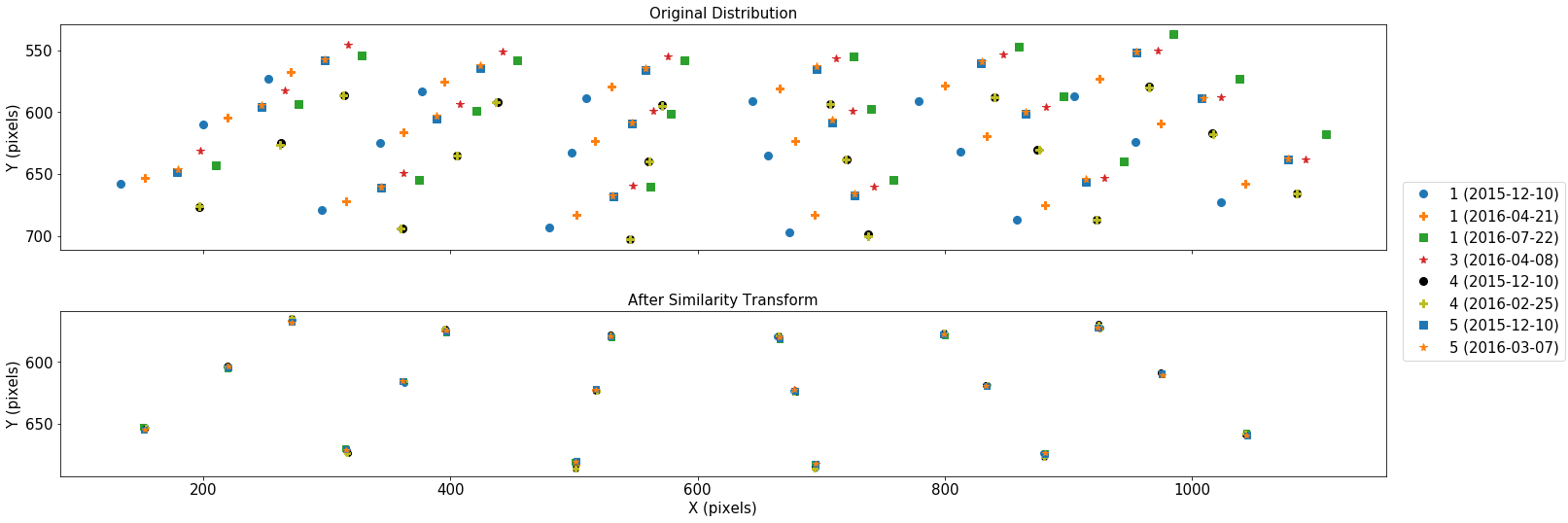}
    \caption{Distribution of Calibration points across setups, before (top) and after (bottom) calibration.}
    \label{FIG_RIG_TRANSFORM}
\end{figure}

Since the data was captured using one of four different rigs, a calibration routine was required to bring all videos in the same frame of reference, simplifying downstream processing.
We were given access to a number of calibration videos for each rig at multiple points throughout the data collection process (but always in between whole recordings), yielding 8 distinct configurations.
Using a checkerboard pattern on the bottom of the rig (see Fig.\ 4(b) in main paper), we annotated 18 reproducible points, which we split into training (odd) and validation (even) sets.
Note that the chosen points correspond to the centres of the \gls{rfid} antennas on the baseplate, but this is immaterial to this calibration routine.
We then used the training set of points to fit affine transforms of varying complexity and also optimised which setup to use as the prototype.
We found that a Similarity transform proved ideal in terms of residuals on the validation set, as seen in Fig.\ \ref{FIG_RIG_TRANSFORM}.

\subsection{Annotation schema}
\label{SS_ANNOT_SCHEMA}
We annotated our data using the VIA tool \cite{MISC_054}.
The annotation requires drawing axis-aligned \glspl{bb} around the mice and the tunnel: the latter was included since we expect future work to incorporate the presence of the tunnel in reasoning about occlusion.
For each frame, we annotate all visible objects (mice/tunnel): hidden objects are defined by their missing an annotation.
For the mice we do not include the tail or the paws in the \gls{bb}: the tunnel includes the opening.
For each annotation, we specify:
\begin{itemize}
\item The identity of the mouse as Red/Green/Blue: we allow specifying multiple tentative identities when it cannot be ascertained with certainty, or when the mice are huddling and immobile (in the latter case, the \gls{bb} extends over the huddle).
\item The level of occlusion: Clear, Truncated or NA when it is difficult to judge/marginal. Note that for this purpose, truncation refers to the size of the \gls{bb} relative to how it would like like if the mouse were not occluded. In other words, it may be possible for a mouse to be partially occluded in such a way that the \gls{bb} is not Truncated.
\item For detections which are hard to make out even for a human annotator, \eg when the mouse is almost fully occluded, a `Difficult' flag is set.
\end{itemize}
Our policy was to annotate frames at 4s intervals.
Within each 30-minute segment, we annotate three snippets, at 00:00 -- 03:00, 13:32 -- 16:28 and 27:00 -- 29:56.
The last snippet does not include 30:00, since the last recorded frame is just before the 30:00 minute mark.
In some segments, we alternatively/also annotate every minute on the minute, again with the caveat that the last annotation is at 29:56 to follow the 4s rule.

\section{Theory}

\subsection{\acrshort{ilp} formulation as covering problem}
\label{SS_ILP_PROOF}
We show that our \gls{ilp} formulation is a special case of the covering problem \parencite{MTH_007}.

The general cover problem is defined as a linear program:
\begin{align}
\text{minimise}   & \quad \sum_{e} c_e x_e , \label{EQ_LP_OBJ}\\
\text{subject to} & \quad \sum_{e} d_{fe} x_e \geq b_f \ \forall \ f \label{EQ_LP_CONSTRAINT}\\
                  & \quad x_e \in \left\lbrace 0, 1 \right\rbrace
\end{align}
in which, the constraint coefficients $d$, the objective costs $c$ and the constraint values $b$ are \emph{non-negative} \cite[p.~109]{MTH_007}.
Here $x_e$ represents an assignment variable and $f$ iterates over constraints.

Let 
\begin{equation}
    \mathbf{x} \equiv \left[\mathbf{a}_{0}\Transpose, \dots , \mathbf{a}_{I+T}\Transpose\right]\Transpose ,
\end{equation}
represent the flattened assignment matrix $A$, where $\mathbf{a}_i$ are the rows of the original matrix.
Index $e$ is consequently the index over $i/j$ in row-major order given by:
\begin{equation}
e = (i-1) (J+1) + j \quad \forall\ i \in I+T,\ j \in J+1 .
\end{equation}
If we assume that log-probabilities are always negative, our original objective, Eq.\ (1) involves \emph{maximising} a sum of \emph{negative} weights, which is equivalent to minimising the sum of the negated weights, and thus would fit the condition that the $c_e$'s in Eq.\ \eqref{EQ_LP_OBJ} should be non-negative.
In reality, probability densities may be greater than 1, and hence our assumption that the weight is always negative may not hold.
However, we can circumvent this if we note that after computing all the weights, we can subtract a constant term such that all weights are negative.
This means we can define $c$ as:
\begin{equation}
    \mathbf{c} \equiv \left[-\mathbf{w}_{0} \Transpose + k\mathbf{1} , \dots , -\mathbf{w}_{I+T}\Transpose + k\mathbf{1}\right]\Transpose ,
\end{equation}
where $\mathbf{w}_i$ are again the rows of the original weight matrix.
In this case, $k$ would be the maximum value over $W$, added to all the elements, such that $\min c = 0$.
Note that in our implementation we do not need to do this, since the solver is not impacted by the formulation as a covering problem.

As for the constraints of our original formulation, Eqs.\ (2 -- 4), these are all summations over the assignment matrix (now represented by $x$) and the coefficients $d_{fe} \in \left\lbrace0,1\right\rbrace$, encoding over which entries (in $x$) we perform summation (for \eg the set of visible tracklets active at time $t$).
These are clearly non-negative as required in the covering problem formulation.

\subsection{Motivation for Performance given detections}
\begin{figure}[!ht]
	\centering
	\includegraphics[width=0.9\textwidth]{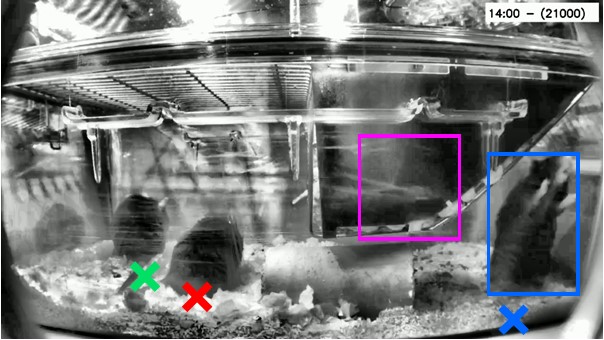}
  	\caption{Hypothetical scenario in which the performance of the Identifier is overshadowed by the detector (only Blue ground-truth is shown to reduce clutter).}
  	\label{FIG_JUSTIFICATION_GD}
\end{figure}

Consider the scenario in Fig.\ \ref{FIG_JUSTIFICATION_GD}.
The ground-truth assignment for the `Blue' mouse is shown as a blue \gls{bb}: however, the detector only outputs the \gls{bb} in magenta.
Clearly, the right thing for the identifier to do is to not assign the magenta \gls{bb} to any mouse, especially if the system is to be used down the line to say predict mouse behaviour.
However, under the definition of overall accuracy, whatever the identifier does, its score on this sample will always be 0.

Our metrics conditioned on the oracle assignment are designed to handle such cases, and tease out the performance of the Identifier from any shortcomings of the detector.
Note how in the same example, the oracle assignment will reject the magenta \gls{bb} as having no identity: hence, the identifier can now get a perfect score if it correctly rejects the \gls{bb}.
This is possible because of the key difference between $A_{O}$ and $A_{GD}$: the former evaluates how well objects $j$ are identified, but the latter measures identification performance for detections $i$.

\section{Fine-Tuning}
\label{APP_TUNING}

In this section, we comment on the various details of the methods employed, including how we came to chose certain configurations over others.

\subsection{Finetuning the Detector}
\label{SS_FINETUNE_DETECTOR}
We use the network pre-trained on CoCo \cite{DS_015} with the typical ``1x'' training settings (see public codebase \footnote{https://github.com/tianzhi0549/FCOS}).
For this, our grayscale images are replicated across the colour channels to yield RGB input to the model.
We remove the \textit{COCO} classifier layer of 81 classes and set a randomly initialised classifier with three categories (\ie,~\textit{mouse, tunnel and background}), and proceed to optimise the weights of the entire network.
We investigate training using either:
\begin{enumerate*}[label=(\alph*)]
    \item the full set of 3247 annotated frames (\glspl{bb} without identities), \textsl{or}
    \item \label{IT_REDUCED_SUBSET} a reduced subset of 3152 frames, discarding frames marked as Difficult (see \sect \ref{SS_ANNOT_SCHEMA}).
\end{enumerate*}
We set a constant learning rate 0.0001 for 30,000 iterations with batch size 16, checkpointing every 1000 iterations.
Evaluation on the held-out annotations from the validation set, indicated that the best performing model was that trained on the reduced subset \ref{IT_REDUCED_SUBSET}, for 11,000 iterations.

\subsection{Finetuning Tracker parameters}
\label{SS_FINETUNE_TRACKER}
\begin{figure}
	\centering
	\begin{tabular}{>{\centering\arraybackslash}m{0.45\textwidth}@{\hspace{1em}}>{\centering\arraybackslash}m{0.45\textwidth}}
		\includegraphics[width=0.45\textwidth]{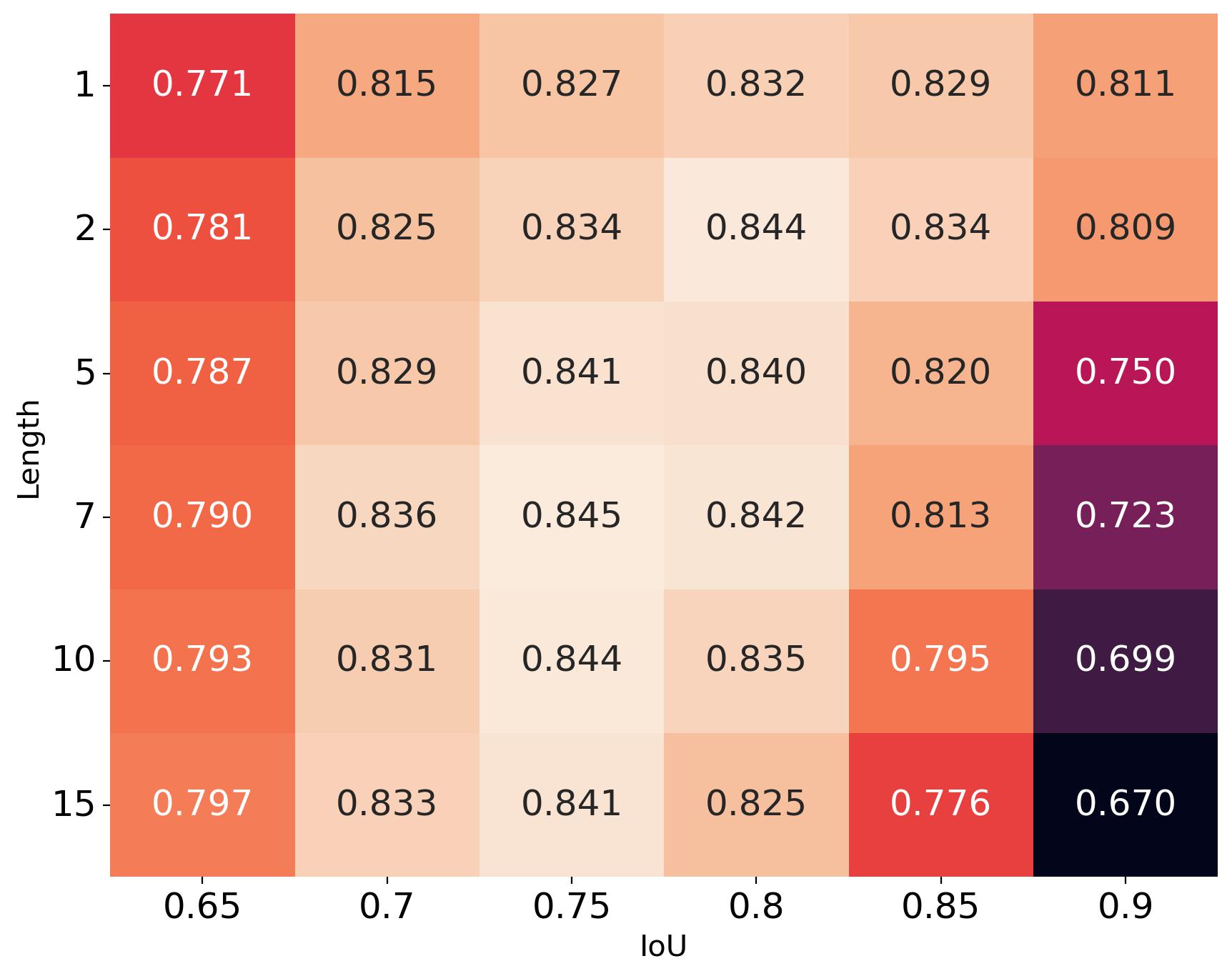} &
		\includegraphics[width=0.45\textwidth]{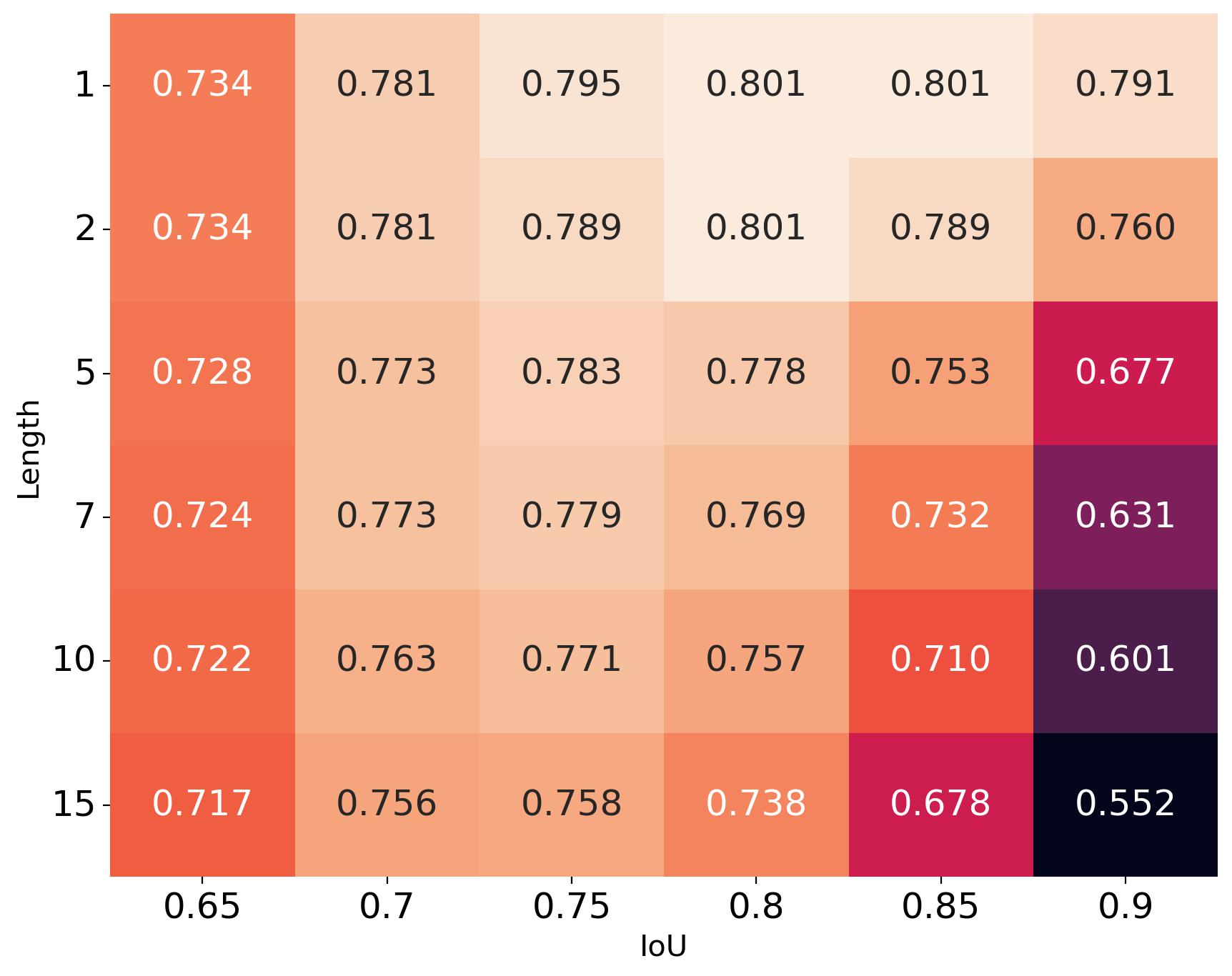} \\
		(a) & (b)
	\end{tabular}
  	\caption{Accuracy (a) given detections, \textsl{and} (b) overall, for various configurations of the tracker \gls{iou} threshold and the minimum tracklet length.}
  	\label{FIG_ACCURACY_GRID}
\end{figure}

Using the combined training/validation set, we ran the identification process using our \gls{ilp} method for different combinations of the parameters, in a grid-search, with \gls{iou} $\in \left\{ 0.65, 0.7, 0.75, 0.8, 0.85, 0.9 \right\}$ and length $\in \left\{1, 2, 5, 7, 10, 15\right\}$.
The resulting accuracies appear in Fig.\ \ref{FIG_ACCURACY_GRID}.
In terms of the metric given detections, (a), the best score is achieved with $L=7$ and
\gls{iou}$=0.75$.
However, we see that the alternative configuration $L = 2$, \gls{iou}$=0.8$ is only marginally worse, and it is also among the optimal in terms of Overall Accuracy (b).
We thus opted for the latter configuration as a compromise between the two scores.

\subsection{Finetuning the Weight model}
Tuning the parameters of the weight model is perhaps the central aspect which governs the performance of our solution. We document this in some depth below.

\paragraph{Choice of Context Vector:}
Using data from the combined training/validation set, and ignoring marginal visibility outcomes, we compared the mutual information between various choices of summarising the configuration $c$ and the annotated visibility, and settled on counting the number of occupancies in each cell as the imparting the most information.
This is natural, but we wanted to see if there was any simpler representation (such as leveraging left/right symmetry).

\paragraph{Predicting $\mathbf{v_j}$:}
Given position $p_j$ and context $c_j$, our next task is to generate a distribution over visibility $v_j$.
We explored using multinomial \gls{lr}, \gls{nb}, decision trees, \glspl{rf} and three-layer \glspl{nn}, all of which we optimised using 5-fold cross-validation on the combined training/validation data.
Note that since we will be summing out $v_j$, getting a calibrated distribution is the key requirement, and hence, when choosing/optimising the models, we used the held-out data log-likelihood as our objective.
Our best score was achieved with a \gls{rf} classifier, with 100 estimators, a maximum depth of 12, minimum samples per-split at 5 and minimum samples at leaf nodes of 2.
Despite the seemingly excessive depth, the mean number of leaf nodes per estimator was around 140 (and never going above 180), due to the regularisation nature of the ensemble method.
The chosen configuration was eventually retrained on the entire training/validation dataset and used in the final models.

\paragraph{Probability of \glspl{bb} under the object model:}
As in Eq.\ (5) from the main paper, when the object is presumed hidden ($v_j = $ Hidden), the probability mass function is a Kronecker delta with probability 0 for all detections, and 1 for the hidden \gls{bb}.
It remains to define the parameters of the Gaussian distribution over visible \glspl{bb} when $v_j$ is Clear or Truncated, which we fit to the training/validation set, but discarding annotations marked as marginal.

We start with defining the means.
In the case of the \gls{bb} centroids, $(x,y)$, we verified that the mean values per-antenna pickup were largely similar across clear/truncated visibilities.
Subsequently, we fit a single model per-antenna.
Given that there is limited data for some antennas, we sought to pool information from the geometrical arrangement of neighbouring antennas.
To this end, we fit a  Homography \cite{MTH_015} between the antenna positions on the baseplate (in world-coordinates, as they appear in Fig.\ 4(b) in main body of the paper), and the corresponding \gls{bb} annotations (in image-space).
The homography construct provides a principled way of sharing statistical strength across the planar arrangement.

On the other hand, inline with our intuition, a Hotelling T2 test \cite{MTH_012} on the sizes of the \glspl{bb}, $(w,h)$, yielded significant differences (with an 0.01 significance level) between \glspl{bb} marked as clear or truncated in all but three of the antenna positions.
Subsequently, we fit individual models per-visibility per-antenna.
Again, we find the need to pool samples for statistical strength, and again we leverage the geometry of the cage.
Specifically, our belief is that the sizes of the \glspl{bb} will vary mostly with depth (into the cage) due to perspective, and are largely stable across antennas at the same depth (or conceptual row, in Fig.\ 4(b)).
We thus estimate the mean sizes on a per-`row' basis, such that for example, antennas 1, 4, 7, 10, 13 and 16 share the same mean size.

We next discuss estimating the covariance.
In actuality, fitting both the Homography on the centroids and the size-means on a per-row/per-visibility basis assumes uniform variance in all directions.
However, we investigated computing a more general covariance matrix using the centred data given the `means' inferred from the above step.
We explored three scenarios:
\begin{enumerate*}[label=(\alph*)]
    \item uniform variance on all dimensions (spherical, as assumed by our mean models),
    \item \label{IT_COV_POOLED} full covariance, pooled across antennas and visibilities, \textsl{and}
    \item \label{IT_COV_PER_VIS} full covariance pooled across antennas but independently per visibility.
\end{enumerate*}
We fit the statistics on the training set and compared the log-likelihood of the validation-set annotations: Table \ref{TAB_COV_MODELS} shows that option \ref{IT_COV_PER_VIS} achieved the best scores and was chosen for our weight model.

\begin{table}
    \centering
    \begin{tabular}{lccc}
	\toprule
                    & Spherical & Pooled Covariance &  Per-Visibility \\
    \cmidrule(lr){2-4}
    Training LL     & -21.97    & -21.35            &  -21.30		  \\
    Validation LL   & -21.83    & \textbf{-21.39} 	&  -21.44		  \\
    \bottomrule
	\end{tabular}
    \caption{Average Log-Likelihoods for Gaussian distribution of \gls{bb} geometries under different choices of the covariance matrix}
    \label{TAB_COV_MODELS}
\end{table}

\paragraph{Probability of \glspl{bb} under the outlier model:}
For the general outlier model we fit a Gaussian independently on the centroid, $(x,y)$ and size, $(w,h)$ components.
Ideally, we would use a uniform distribution over the size of the image for the former, but in the interest of uniformity with the size-model, we approximate it with a broad Gaussian.
The parameters for the sizes are simply estimated on all \gls{bb} annotations in our training/validation data.
\section{Description of the Supporting Material}
\label{APP_SUPP}

We provide the Dataset, Code and Sample Video Clips at \url{https://github.com/michael-camilleri/TIDe}

\subsection{DataSet}
We have packaged a curated version of the dataset, containing the annotations, detections and position pickups as pandas dataframes.
Further details appear in the repository's Readme.

\subsection{Code}
We make available the tracking, identification and evaluation implementations: note that the code is provided as is for research purposes and scrutiny of the algorithm, but is not guaranteed to work without the accompanying setup/frameworks.
\begin{itemize}
\item `Evaluation.py': implements the various evaluation metrics,
\item `Trackers.py': implementing our changes to the SORT Tracker and packaging it for use with our code,
\item `Identifiers.py': implementing the Identification problem and its solution.
\end{itemize}

\subsection{Dataset Sample}

\paragraph{\href{https://github.com/michael-camilleri/TIDe/blob/main/TIM/results/Clip_1.mp4}{Clip 1}:}
This shows a sample video-clip of 30s from our data.
The original raw video appears on the left: on the right, we show the same video after being processed with CLAHE \cite{MISC_038} to make it more visible to human viewers, together with the \gls{rfid}-based position of each mouse overlayed as coloured dots.
Note that our methods all operate on the raw video and not on the CLAHE-filtered ones.

\subsection{Sample Tracking and Identification Videos}
\label{SS_VIDEOS}
We present some illustrative video examples showing the \gls{bb} tracking each mouse as given by our identification method, and comparisons with other techniques.
Apart from the first clip, the videos are played at 5 \gls{fps} for clarity, and illustrate the four methods we are comparing in a four-window arrangement with Ours on the top-left, the baseline on the top-right, the static assignment with the probabilistic model bottom-left and the SiamMask tracking bottom-right.
The frame number is displayed in the top-right corner of each sub-window.
The videos are best viewed by manually stepping through the frames.

\paragraph{\href{https://github.com/michael-camilleri/TIDe/blob/main/TIM/results/Clip_2.mp4}{Clip 2}:} % 4154665_139_(42375-43875)
We start by showing a minute-long clip of our identifier in action, played at the nominal framerate of 25\gls{fps}.
This clip illustrates the difficulty of our setup, with the mice interacting with each other in very close proximity (\eg Blue and Green in frames 42375 -- 42570) and often partially/fully occluded by cage elements such as the hopper (\eg Red in frames 42430 -- 42500) or the tunnel (\eg Blue in frames 42990 -- 43020).

\paragraph{\href{https://github.com/michael-camilleri/TIDe/blob/main/TIM/results/Clip_3.mp4}{Clip 3}:} % 4184108_92_(40575-40650)
This clip shows the ability of our model to handle occlusion.
Note for example how in frames 40594 -- 40605, the relative arrangement of the \gls{rfid} tags confuses the baseline (top-right), but the (static) probabilistic weight model is sufficient to reason about the occlusion dynamics and the three mice.
Temporal continuity does however add an advantage, as in the subsequent frames, 40607 -- 40630, even the static assignment (bottom-left) mis-identifies the mice, mostly due to the lag in the \gls{rfid} signal.
The SiamMask (bottom-right) fails to consistently track any of the mice, mostly because of occlusion and passage through tunnel which happens later on in the video (not shown here), and shows the need for reasoning about occlusions.

\paragraph{\href{https://github.com/michael-camilleri/TIDe/blob/main/TIM/results/Clip_4.mp4}{Clip 4}:} % 4184108_92_(40850-40900)
This clip shows the weight model successfully filtering out spurious detections.
For clarity, we show only the static assignment (left) and the baseline (right).
Note how due to a change in the \gls{rfid} antenna for Green, its \gls{bb} often gets confused for noisy detections below the hopper (see \eg frames 40867 -- 40875): however, the weight model, and especially the outlier distribution, is able to reject these and assign the correct \gls{bb}.

\paragraph{\href{https://github.com/michael-camilleri/TIDe/blob/main/TIM/results/Clip_5.mp4}{Clip 5}:} % 4184108_92_(40975-41025)
This shows another difficult case involving mice interacting and hiding each other.
The SiamMask is unable to keep track of any of the mice consistently.
While our method does occasionally lose the red mouse when it is severely occluded (\eg frame 40990) the baseline gets it completely wrong (mis-identifying green for red in \eg frames 40990 -- 41008), mostly due to a lag in the \gls{rfid} which also trips the static assignment with our weight model.

\paragraph{\href{https://github.com/michael-camilleri/TIDe/blob/main/TIM/results/Clip_6.mp4}{Clip 6}:} % 4181380_19_(40650-40700)
Finally, this shows an interesting scenario comparing our method (left) to the SiamMask tracker (right).
The latter is certainly smoother in terms of the flow of \glspl{bb}, with the tracking-by-detection approach understandably being more jerky.
However, SiamMask's inability to handle occlusion comes to the fore towards the end of the clip (frames 40693 -- end), where the green track latches to the face of the blue mouse.

%------------------------------------------------------%

\clearpage
\bibliographystyle{IEEEtranN}
%\bibliography{../../../../../Literature/library}
% Generated by IEEEtranN.bst, version: 1.13 (2008/09/30)

%------------------------------------------------------%

\end{document}